\documentclass{article}

\usepackage{microtype}
\usepackage{graphicx}
\usepackage{subfigure}
\usepackage{booktabs} 
\usepackage{caption} 
\usepackage{float}
\usepackage{chngcntr}
\usepackage{multirow} 
\usepackage[linesnumbered,ruled,vlined]{algorithm2e}

\usepackage{hyperref}

\usepackage[accepted]{icml2024}

\usepackage{amsmath}
\usepackage{amssymb}
\usepackage{mathtools}
\usepackage{amsthm}

\usepackage{pifont}
\newcommand{\xmark}{\ding{55}}%

\usepackage[capitalize,noabbrev]{cleveref}

\theoremstyle{plain}

\theoremstyle{definition}

\theoremstyle{remark}

\usepackage[textsize=tiny]{todonotes}

\usepackage{colortbl}
\definecolor{ourbg}{rgb}{0.9, 0.9, 1.}  
\definecolor{darkgray}{rgb}{0.43,0.43,0.43}  
\definecolor{grayarea}{rgb}{0.95, 0.95, 0.96}
\definecolor{llm}{rgb}{0.97, 0.97, 0.93}
\definecolor{ouralg}{rgb}{0.8, 0.8, 0.9}

\usepackage{amsmath,amsfonts,bm}

\def\1{\bm{1}}

\def\vm{{\bm{m}}}

\def\vy{{\bm{y}}}
\def\vz{{\bm{z}}}

\def\mE{{\bm{E}}}

\DeclareMathAlphabet{\mathsfit}{\encodingdefault}{\sfdefault}{m}{sl}
\SetMathAlphabet{\mathsfit}{bold}{\encodingdefault}{\sfdefault}{bx}{n}

\newcommand{\x}{$\times$}

\DeclareMathOperator*{\argmax}{arg\,max}

\newcommand{\ourmethod}{MC-ViT}
\renewcommand{\x}{\times}

\usepackage{xspace}

\makeatletter
\DeclareRobustCommand\onedot{\futurelet\@let@token\@onedot}
\def\@onedot{\ifx\@let@token.\else.\null\fi\xspace}

\def\eg{\emph{e.g}\onedot} 
\def\ie{\emph{i.e}\onedot}

\makeatother

\usepackage{algorithm}
\usepackage{listings}

\usepackage{etoolbox}
\makeatletter
\AfterEndEnvironment{algorithm}{\let\@algcomment\relax}
\AtEndEnvironment{algorithm}{\kern2pt\hrule\relax\vskip3pt\@algcomment}
\let\@algcomment\relax
\newcommand\algcomment[1]{\def\@algcomment{\footnotesize#1}}
\renewcommand\fs@ruled{\def\@fs@cfont{\bfseries}\let\@fs@capt\floatc@ruled
  \def\@fs@pre{\hrule height.8pt depth0pt \kern2pt}%
  \def\@fs@post{}%
  \def\@fs@mid{\kern2pt\hrule\kern2pt}%
  \let\@fs@iftopcapt\iftrue}
\makeatother

\begin{document}

\twocolumn[
\icmltitle{Memory Consolidation Enables Long-Context Video Understanding}



\icmlsetsymbol{equal}{*}

\begin{icmlauthorlist}

\icmlauthor{Ivana Bala\v{z}evi\'c}{equal,comp}
\icmlauthor{Yuge Shi}{equal,comp}
\icmlauthor{Pinelopi Papalampidi}{equal,comp} \\
\icmlauthor{Rahma Chaabouni}{comp}
\icmlauthor{Skanda Koppula}{comp}
\icmlauthor{Olivier J. H\'enaff}{comp}

\end{icmlauthorlist}

\icmlaffiliation{comp}{Google DeepMind}

\icmlcorrespondingauthor{Ivana Bala\v{z}evi\'c}{balazevic@google.com}
\icmlcorrespondingauthor{Olivier J. H\'enaff}{henaff@google.com}

\icmlkeywords{Machine Learning, ICML}

\vskip 0.3in
]



\printAffiliationsAndNotice{\icmlEqualContribution} 

\begin{abstract}

Most transformer-based video encoders are limited to short temporal contexts due to their quadratic complexity. While various attempts have been made to extend this context, this has often come at the cost of both conceptual and computational complexity. 
Instead, we propose to re-purpose existing pretrained video transformers by simply fine-tuning them to attend to memories derived non-parametrically from past activations. By leveraging redundancy reduction, our \textit{memory-consolidated vision transformer} (\ourmethod) effortlessly extends its context far into the past and exhibits excellent scaling behavior when learning from longer videos. 
In doing so, \ourmethod \ sets a new state-of-the-art in long-context video understanding on EgoSchema, Perception Test, and Diving48, outperforming methods that benefit from orders of magnitude more parameters. 

\end{abstract}

\begin{figure}[t]
\begin{center}
\centerline{\includegraphics[width=\linewidth]{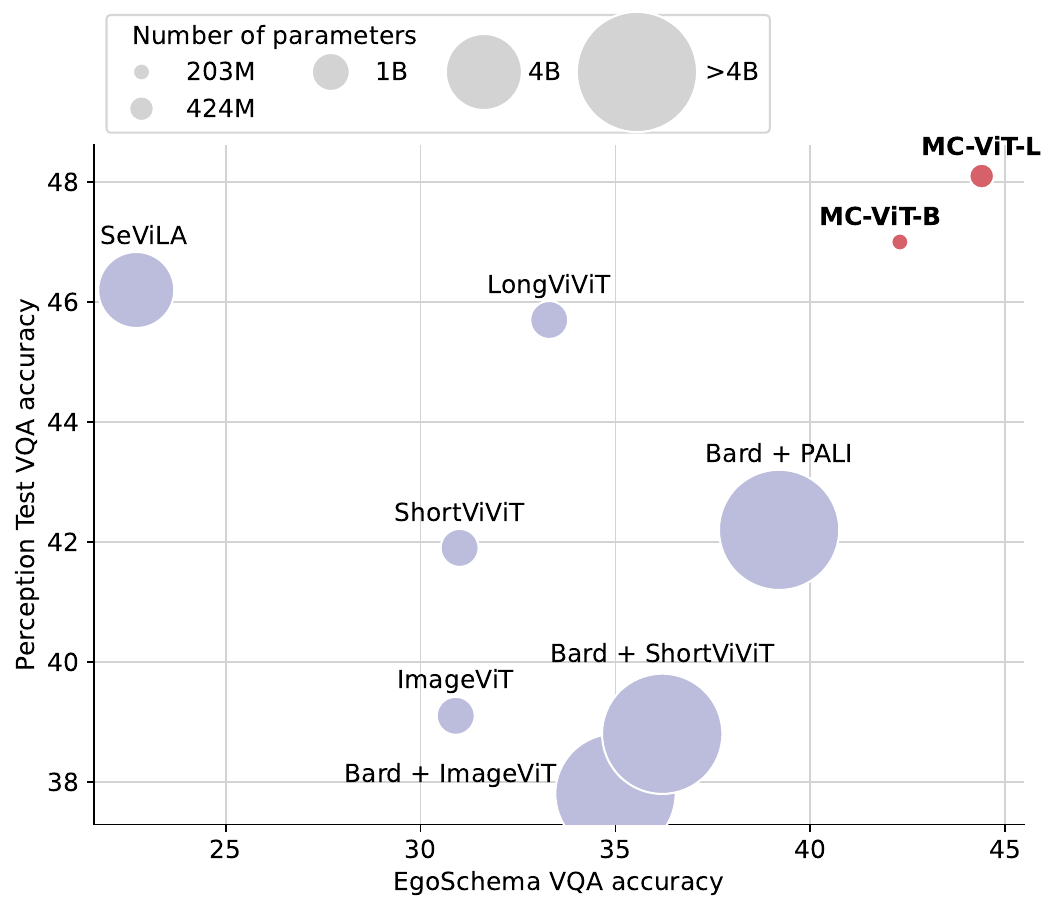}}
\vspace{-0.5em}
\caption{\textbf{Long-context video understanding on EgoSchema and Perception Test.} The proposed Memory-Consolidated Vision Transformer (MC-ViT-\{B,L\}, shown in bold) surpasses both public and large-scale proprietary models, despite using orders of magnitude fewer parameters and requiring only short fine-tuning schedules on top of standard pretrained models.} 
\vspace{-0.5em}
\label{fig:teaser}
\end{center}
\end{figure}

\section{Introduction}

\label{sec:intro}

Humans and animals reason about events extending over days, weeks, and years \cite{tulving1985memory}, yet current artificial vision systems live largely in the present. While architectures that model the dynamics of natural videos have grown ever more sophisticated \cite{carreira2017quo,feichtenhofer2019slowfast,arnab2021vivit}, the temporal extent over which they reason has typically been limited to a small number of frames. In particular, transformer architectures \cite{vaswani2017attention} which power most applications in vision and language do not scale to the vast number of tokens present in natural videos due to their quadratic complexity.
For example, 30 minutes of video sampled at standard rates may contain half a million tokens--- more than what current state-of-the-art architectures using optimized attention algorithms~\citep[\eg][]{dao2022flashattention} can process. 
Several attempts have been made to extend the temporal context of video transformers, including masking, attention approximations, and parametric memory modules \citep[\eg][]{wu2022memvit, piergiovanni2023mirasol3b}. However, these approaches often introduce additional complexity, requiring
specialized architectures and training paradigms.

In this work, we question whether such modifications are indeed necessary to enable long-context modeling. Starting from standard pretrained video transformers \cite{arnab2021vivit}, we process videos in a streaming setting in order to bound their complexity by the length of short segments \cite{dai2019transformer}. Crucially, we process individual segments in relation to a memory bank which is populated non-parametrically with the consolidated activations from past segments. This allows us to re-purpose pretrained video transformers for long-context understanding without any architectural modification, by simply fine-tuning them to attend to this memory with short training schedules. 

A central question we are faced with is therefore how to choose which of the quasi-infinite tokens from past frames to store in memory. Inspired by evidence from psychology and neuroscience which formulates memory as a reconstructive process \cite{bartlett1932remembering,marr1971simple,spens2024generative}, we adopt simple nonparametric schemes that form memories that are maximally representative of the full set of past activations. We find these mechanisms to effectively compress memories by an order of magnitude, and allow our \textit{memory-consolidated vision transformer} (\ourmethod) to extend its context to significantly longer videos while maintaining a bounded complexity.
In particular, 

\hspace{1em} 1.\ \ourmethod \ strikes a favorable trade-off between computational complexity and expressivity, outperforming standard video transformers and efficient approximations thereof with 10$\x$ less memory and computation.

\hspace{1em} 2.\ The non-parametric nature of \ourmethod \ allows us to straightforwardly re-purpose off-the-shelf pretrained video transformers by fine-tuning them to use their consolidated memory, yielding large efficiency gains by decreasing overall training time on long videos. 

\hspace{1em} 3.\ \ourmethod \ sets a new state-of-the-art on long-context video understanding tasks such as fine-grained action recognition (Diving48) and video question answering (EgoSchema and Perception Test), outperforming methods which benefit from orders of magnitude more parameters.

\hspace{1em} 4.\ \ourmethod \ is competitive with large-scale proprietary systems such as GPT-4V and Bard, despite using a small, standard, and open architecture and training paradigm.

\begin{figure*}[h]
\centering
\includegraphics[width=\linewidth]{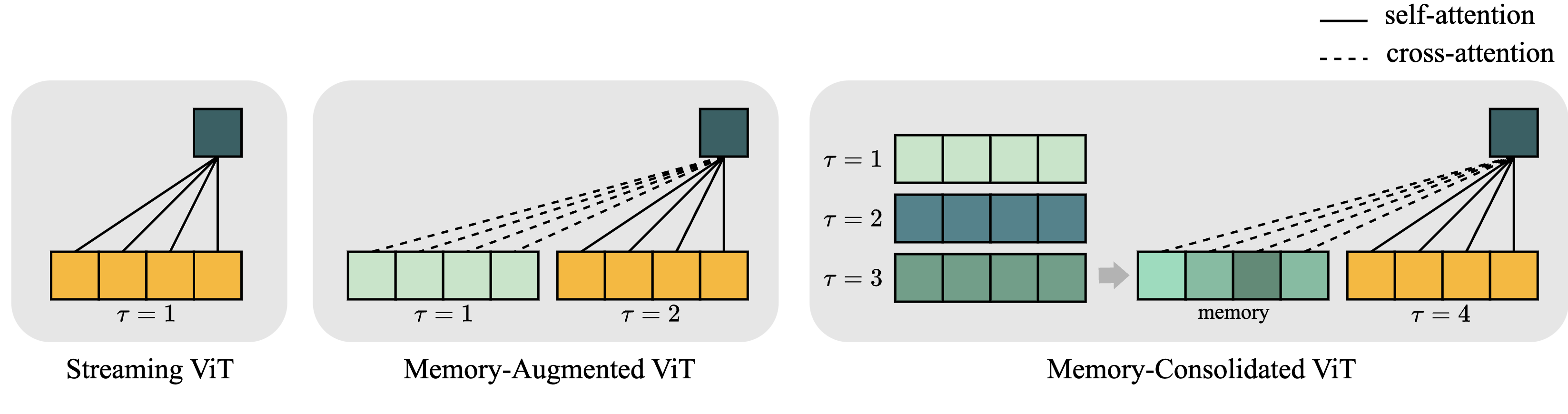}
\vspace{-3em}
\caption{\textbf{Visualization of the proposed method.} \textbf{Left: } Streaming ViT processes each segment of the sequence independently by attending over activations within a segment. \textbf{Middle:} Memory-Augmented ViT, similar to Transformer XL \citep{dai2019transformer}, attends to current activations (yellow blocks) and those in recent history (green blocks). \textbf{Right: } In Memory-consolidated ViT, we consolidate the extended context into shorter memory and cross-attend over them, which enables us to effectively attend over longer sequences. \vspace{-1em}}\label{fig:visualization_three}
\end{figure*}

\section{Related Work}
\label{sec:rw}

\noindent \textbf{Long-context architectures.} 
Prior work has thoroughly explored approaches for handling long textual or visual inputs, by sparsifying either the input tokens or the attention applied over these tokens. In natural language processing, notable examples include Big Bird~\cite{zaheer2020big} and LongFormer~\cite{Beltagy2020Longformer} that employ local self-attention over restricted windows combined with global tokens that attend over the entire sequence. Alternative attention mechanisms in vision have utilized pooling~\citep{wang2021pyramid,li2022mvitv2}, linear~\citep{bolya2022hydra} and windowed formulations~\citep{dong2022cswin,li2022exploring,ryali2023hiera}. 
Several works reduce the number of tokens via multi-resolution patchification, thus processing the input video at different granularities~\citep{feichtenhofer2019slowfast,yan2022multiview,piergiovanni2023rethinking}. Similarly, \citet{papalampidi2023simple} showcase the benefits of this approach by training video encoders on long contexts with high ratios of input masking. 
Current state-of-the-art approaches for processing long videos consist of modular systems for captioning and extracting frame-level information, followed by a billion-scale LLM for aggregating this information~\cite{zeng2022socratic,wang2022language,li2023videochat,lin2023mmvid,wang2023vamos,zhang2023simple}. 
The approach proposed in this work is orthogonal to these, by re-purposing standard transformer architectures for long-context modeling, whose representations can be incorporated into LLMs.

\noindent \textbf{Memory-augmented transformers.} Since the introduction of transformers \cite{vaswani2017attention}, several works have sought to give them additional context via auxiliary memory banks. In NLP, TransformerXL does so by simply attending to recent activations in a streaming setting \cite{dai2019transformer}, whereas Retro \cite{borgeaud2022improving} does so by retrieving semantically related content. In vision, memory-augmented architectures have also been shown to enable video object segmentation \cite{oh2019video}, tracking \cite{lai2020mast}, and action recognition~\cite{wu2019long}. 

\noindent \textbf{Memory-compressing transformers.} Several transformer-based architectures explored compressing past activations into a finite-length memory. In NLP, Neural Turing Machines \cite{graves2014neural} and Token Turning Machines \cite{ryoo2023token} learn to read and write from a memory bank in an end-to-end manner. Similarly, Compressive Transformers \cite{rae2019compressive}, $\infty$-former~\cite{martins2022former}---and in vision, MemDPC \citep{han2020memory}, LSTR \citep{xu2021long}, MeMViT \citep{wu2022memvit} and LongMem \citep{wang2023augmenting}---extend the effective context length by compressing prior activations with additional parametric modules. Concurrent work Mirasol3B \cite{piergiovanni2023mirasol3b} showcases the power of this approach by combining these memory modules with large language models and a bespoke pretraining protocol.

Our work differs from these in that we find that a simple, non-parametric mechanism followed by light-weight fine-tuning is sufficient to re-purpose standard pretrained video transformer architectures \citep[\eg ViViT,][]{arnab2021vivit} to achieve strong long-context modeling. 
Closest to our approach is concurrent work MovieChat~\cite{song2023moviechat} that also uses a non-parametric memory, but focuses on extending the visual context of an LLM for video-to-text generation while using fixed frame-level representations via a pre-trained image encoder. In contrast, we directly add memory consolidation to the visual backbone.

\section{Method}
\label{sec:method}

\subsection{Overview of Video Vision Transformers (ViViT)} \label{sec:vivit_overview}

Video Vision Transformers (ViViT;~\citealt{arnab2021vivit}) adapt Vision Transformers \cite{dosovitskiy2020image} 
to straightforwardly process videos. Specifically, ViViT divides a video $V \!\in\! \mathbb{R}^{T \times H \times W}$ into $N_T$ non-overlapping spatio-temporal patches $x_i \!\in\! \mathbb{R}^{t \times h \times w}$ such that $N_T \!=\! \frac{T}{t} \cdot \frac{H}{h} \cdot \frac{W}{w}$, and linearly projects these patches into 1D embedding space:
\begin{align}
    z_i = \mE x_i + p_i, \label{eq:vivit_projection}
\end{align}
where $\mE$ denotes a learnable projection layer and $p_i \!\in\! \mathbb{R}^{d}$  additional positional embeddings. The resulting token sequence
$\vz^0 = [z_i, i \!\in\! [1, N_T]] \!\in\! \mathbb{R}^{N_T \times d}$ is then passed through a series of $L$ transformer layers, which alternate Multi-head Self-Attention (MSA;~\citealt{vaswani2017attention}), layer normalization (LN;~\citealt{ba2016layernorm}) and MLP blocks:
\begin{align}
\vy^l & = \textrm{MSA}(\textrm{LN}(\vz^l)) + \vz^l \\ 
\vz^{l+1} & = \textrm{MLP}(\textrm{LN}(\vy^l)) + \vy^l. 
\end{align}
While various schemes for factorizing the attention have been proposed for ViViT, we build our model upon the simplest joint space-time attention which models dependencies between all tokens in the sequence. We leave exploring other factorization methods for future research. In contrast to ViViT's self-attention which spans the entire video, our \ourmethod \ model uses self-attention within much shorter segments, and cross-attention across segments via a set of consolidated memories, which we detail below.  

\vspace{-0.7em}
\subsection{Memory-Consolidated Vision Transformers}
\label{sec:mc-vit}

In this section, we explore three successive modifications to the original ViViT architecture that enable efficient and expressive scaling to longer videos (see visualization in \Cref{fig:visualization_three}). The culmination of these modifications represents our proposed method: 
Memory-Consolidated ViT (\ourmethod). We apply consistent pre- and post-processing steps across all three approaches: we divide the video $V$ into $s$ temporal segments $v_\tau \!\in\! \mathbb{R}^{S \times H \times W}$, where $S = \frac{T}{s}$ is the number of frames per segment and $S\!=\!16$ in our experiments. We then process each segment (either individually or jointly, see below), yielding a list of $s$ representations $\{\vz_1, \cdots, \vz_s\}$, one for each segment. All of these are then concatenated as the final representation of the video.

\textbf{Streaming ViT (ST-ViT).} Since the computational complexity of transformers scales quadratically with the number of tokens, full joint space-time attention becomes intractable for video lengths that exceed even small numbers of frames. To counteract this, we start with a simple streaming-based extension of ViViT, which processes each segment $v_\tau, \tau \!\in\! [1, s]$ independently, as described in \Cref{sec:vivit_overview}, with positional embeddings spanning the entire video. Crucially, the number of tokens processed by the ViViT encoder at a given time is instead $N = \frac{S}{t} \cdot \frac{H}{h} \cdot \frac{W}{w}$, bounding the quadratic complexity by the segment length $S$ rather than the total video length $T$. We include the pseudocode for the streaming ViT implementation in \Cref{sec:app-pseudocode}, \Cref{alg:streaming_vit_algorithm}.

\textbf{Memory-Augmented ViT (MA-ViT).} While more scalable, the streaming setting limits the encoder's ability to reason over events which span multiple segments. Hence, as in \citet{dai2017good}, we augment the self-attention module with an additional set of memories $\vm^l_\tau = [\vz^l_0; \vz^l_1; ...;\vz^l_{\tau-1}] \in \mathbb{R}^{M \times d}$ consisting of concatenated activations of previous segments at each layer $l$: 
\begin{equation}
\label{eq:mca}
\vy^l_\tau  = \textrm{MCA}(\underbrace{\textrm{LN}(\vz^l_\tau)}_{\text{query}}, \underbrace{[\textrm{LN}(\vz^l_\tau); \textrm{LN}(\vm^l_\tau)]}_{\text{key-value}}) + \vz^l_\tau,
\end{equation}
where $[\cdot;\cdot]$ denotes the concatenation operation and Multi-head Cross-Attention (MCA;~\citealt{dai2019transformer}) generalizes MSA by decoupling the inputs to the query and key/value heads. Specifically, the MCA operation allows activations from the current segment $\vz^l_\tau$ to attend both to themselves (as in MSA) and to memories of all past activations $\vm^l_\tau$, while keeping the quadratic complexity limited to $N+M$. We include the pseudocode for Memory-Augmented ViT in \Cref{sec:app-pseudocode}, \Cref{alg:mem_aug_vit_algorithm}.

\begin{algorithm}[t]
\caption{Memory-consolidated ViT.}
\label{alg:mcvit_algorithm}
\definecolor{codeblue}{rgb}{0.25,0.5,0.5}
\lstset{
  backgroundcolor=\color{ouralg!20},
  basicstyle=\fontsize{7.2pt}{7.2pt}\ttfamily\selectfont,
  columns=fullflexible,
  breaklines=true,
  captionpos=b,
  commentstyle=\fontsize{7.2pt}{7.2pt}\color{codeblue},
  keywordstyle=\fontsize{7.2pt}{7.2pt},
}
\begin{lstlisting}[language=python]
def mc_vit(
        video, n_chunks, n_layers, 
        pos_emb, mc_method, num_mem
):
  emb = linear_proj(video) + pos_emb   # [B, N, D]
  chunked_video = np.split(emb, n_chunks, axis=1)
  memory = {layer: None for layer in range(n_layers)}
  zs = []
  for z in chunked_video:
    z_norm = layer_norm(z)
    for layer in range(n_layers):
        if memory[layer] is None:
            y = self_attention(z_norm) + z
        else:
            kv = np.concatenate(z_norm, memory[layer]))
            y = cross_attention(q=z_norm, kv=kv) + z
        y_norm = layer_norm(y)
        z = mlp(y_norm) + y
        memory[layer] = memory_consolidation(
            memory[layer], z, num_mem, mc_method)
        memory[layer] = layer_norm(memory[layer])
        zs.append(z)
  return np.concatenate(zs, axis=1)
\end{lstlisting}
\end{algorithm}


\textbf{Memory-Consolidated ViT (\ourmethod).} Given the memory-augmented vision transformer architecture, a central question is how to consolidate the (potentially infinite) activations of previous segments into a finite (and ideally small) set of memories. 
We consider three simple instances of memory consolidation that model memory through a non-parametric reconstructive process. 

To produce a new consolidated memory $\vm_\tau$ for the current segment (dropping the layer index $l$ for concision), we consolidate the set of activations from the preceding segment $\vz_{\tau-1} \!\in\! \mathbb{R}^{N \times d}$ into $\bm{\hat{z}}_{\tau-1}  \!\in\! \mathbb{R}^{K \times d}$ ($K \le N$) and concatenate them to the memories consolidated from all prior segments $\vm_\tau \!=\! [\vm_{\tau-1}, \bm{\hat{z}}_{\tau-1}] \!\in\! \mathbb{R}^{(M+K) \times d}$. The proposed instances of non-parametric memory consolidation differ in their way of computing $\bm{\hat{z}}_{\tau-1}$, which we detail below.

\textbf{MC-ViT-R (random)} is the simplest non-parametric baseline which randomly selects a set of $K$ activations from $\vz_{\tau-1}$ and uses them as the consolidated memory for the preceding segment:
\begin{align}
\bm{\hat{z}}_{\tau-1}^{\text{R}} &= \{ \vz_{\tau-1,k} \ | \ k \in \mathcal{I} \} \in \mathbb{R}^{K \times d},
\label{eq:random}
\end{align}
where $\mathcal{I}\in [1, N]^K$ is a set of $K$ randomly selected indices.

\textbf{MC-ViT-CS (coreset)} constructs a maximally representative set of memories by applying the greedy coreset selection algorithm \citep{agarwal2005geometric} to the activations of the preceding segment $\vz_{\tau-1}$ by iteratively adding the most distant activations to the ones already included in the consolidated memory for that segment. One iteration of the algorithm is defined as:
 \begin{equation}
 k^* = \underset{k \in [1, N]}{\text{arg\,max}} \; \underset{j \in \mathcal{M}^*}{\text{min}} ||\bm{z}_{\tau-1, k}-\bm{z}_{\tau-1, j}||_2^2
 \end{equation}
 \vspace{-0.5em} 
 \begin{equation}
 \mathcal{M}^* \leftarrow \mathcal{M}^* \cup \{k^*\},
 \end{equation}
where $\mathcal{M}^*$ is the set of activation indices chosen to be added to the consolidated memory $\bm{\hat{z}}_{\tau-1}^{\text{CS}}$. 
The greedy coreset selection algorithm is run for $K$ iterations to produce the consolidated memory $\bm{\hat{z}}_{\tau-1}^{\text{CS}} \!\in\! \mathbb{R}^{K \times d}$ for the segment $v_{\tau-1}$. Due to its iterative nature, the coreset selection algorithm becomes increasingly computationally expensive as the size of the segment memory $K \!=\! |\mathcal{M}^*|$ increases.

\textbf{MC-ViT-KM (k-means)} randomly initializes $K$ cluster centroids as $\bm{\hat{z}}_{\tau-1}^{\text{R}}$ (see Equation \ref{eq:random}) and then performs 5 iterations of k-means clustering on all activations of the previous segment $\vz_{\tau-1}$ to compute the updated cluster centroids, which we use as the consolidated memory $\bm{\hat{z}}_{\tau-1}^{\text{KM}} \!\in\! \mathbb{R}^{K \times d}$ for the segment $v_{\tau-1}$.

We include the pseudocode for \ourmethod\ in \Cref{alg:mcvit_algorithm}. 
The newly consolidated memory $\vm_\tau$ is then jointly processed with the current segment activations $\vz_{\tau}$ via MCA, analogously to MA-ViT (see Equation \ref{eq:mca}).

We compare these different consolidation methods in Section~\ref{sec:exp-memory} and find that MC-ViT-KM performs better than the others. Therefore, unless specified otherwise, MC-ViT refers to MC-ViT-KM in the following sections.

\begin{figure*}[t]
\centering
\includegraphics[width=\linewidth]{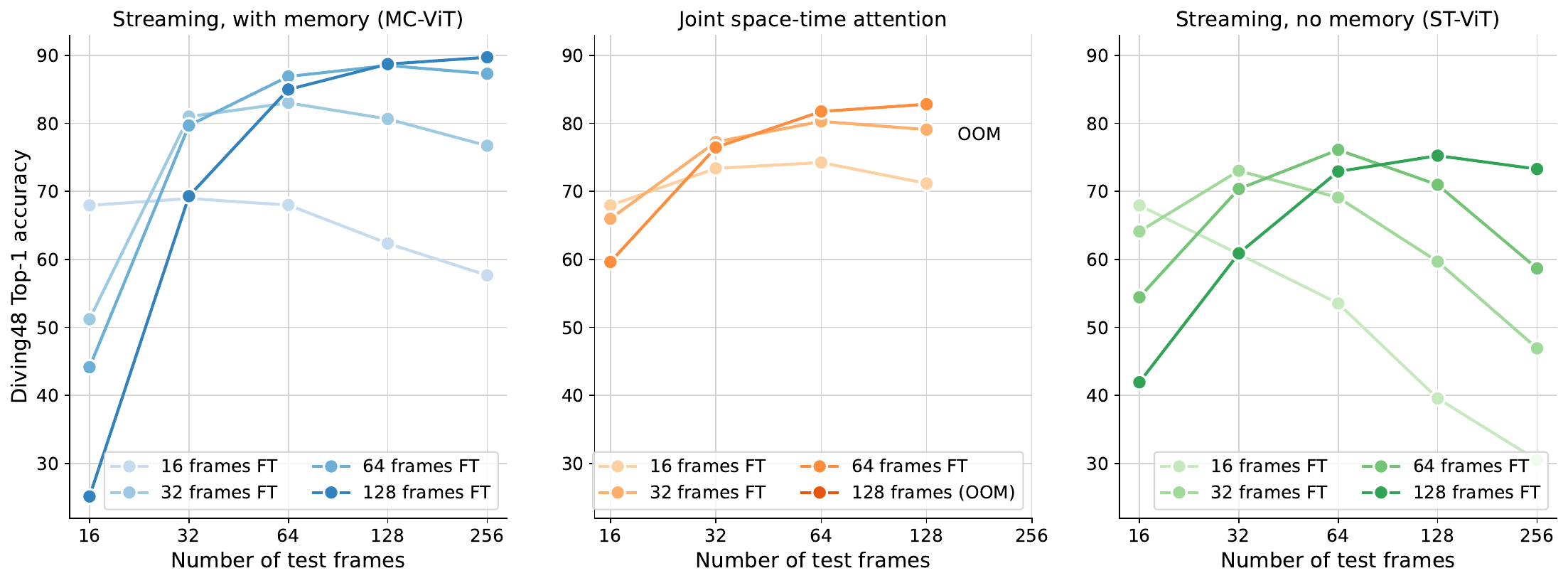}
\caption{\textbf{\ourmethod \ effectively learns from long videos.} \textbf{Left: } \ourmethod \ scales to long Diving48 videos at both training and inference time, and benefits from fine-tuning on longer videos. \textbf{Middle: }Joint space-time attention benefits from fine-tuning on longer videos, but cannot learn from long (128 frame) videos due to its large complexity and memory footprint. \textbf{Right: } ST-ViT scales to longer videos but does not benefit from training on them.\vspace{-1em}}
\label{fig:learning}
\end{figure*}

\subsection{Training and Evaluation}
\label{sec:method-training}

\textbf{Initialization.} Since the parameters of \ourmethod \ are almost identical to those of ViViT \cite{arnab2021vivit}, we initialize most parameters from a ViViT encoder pretrained on short (16-frame) video clips using multimodal contrastive learning \cite{xu2021videoclip,papalampidi2023simple}, see \cref{app:vivit_pretraining}. The only parameters which differ are positional embeddings, as we fine-tune \ourmethod \ on significantly longer videos (\eg up to 128 frames) than the short clips used for pretraining. We therefore initialize the positional embeddings by linearly upsampling ViViT's positional embeddings through interpolation along the time dimension (note that we experimented with both interpolation and extrapolation through repetition and did not find it to have a big impact on performance). Similarly, we re-use and fine-tune a BERT-style language encoder pretrained in the same setup. 

\textbf{Fine-tuning.} For each evaluation, we fine-tune on a dataset mixture that enables a like-for-like comparison with the previous state-of-the-art. All datasets are composed of video-text pairs, and we therefore simply fine-tune the model with noise contrastive estimation. Given the video and text embeddings $\vz^v_i$ and $\vz^t_i$ of an example $i$, we minimize
\begin{equation}
\label{eq:con}
\ell_i = - \log \frac{\exp( \vz^v_i {\cdot} \vz^t_i)}{\sum_j \exp(    \vz^v_i {\cdot} \vz^t_j )}
- \log \frac{\exp( \vz^t_i {\cdot} \vz^v_i)}{\sum_j \exp(    \vz^t_i {\cdot} \vz^v_j )}
\end{equation}
where the ``negative'' embeddings $\vz^v_j$ and $\vz^t_j$ are the in-batch examples unless otherwise specified. We provide further training details in Appendix~\ref{sec:app_training_details}.

\textbf{Evaluation.} We employ the standard zero-shot transfer paradigm from CLIP \cite{radford2021learning} to perform all downstream tasks. In all cases, a test video is equipped with multiple possible ``captions'', only one of which is correct. For action recognition, these captions are simply the class names. For video question answering, captions are question-answer pairs constructed from the set of multiple-choice answers. We utilize the language model to compute caption embeddings $\vz^t_i$, and compare them to the video embedding $\vz^v_i$. The model's prediction $i^* = \argmax_i \vz^v_i {\cdot} \vz^t_i$ is simply the caption with the highest similarity to the video embedding.

\section{Experiments}
\label{sec:exp}

\subsection{Datasets}

We evaluate our method on four challenging datasets for long-context video understanding, namely Diving48, EgoSchema, Next-QA, and Perception Test.

\textbf{Diving48} \citep{li2018diving48} was specifically designed to assess the importance of dynamic and long-term temporal reasoning in action recognition. Video lengths vary between 24 and 822 frames, with 158 frames on average. Each video is categorized into 48 fine-grained classes based on the specific dive type it depicts. Consequently, correct classification requires dense video sampling and fine-grained understanding in addition to retaining information over a long temporal extent, which necessitates reasoning over a large number of frames. To align with prior methods, we fine-tune on the Diving48 training set and re-initialize the language encoder randomly with a linear embedding function.

\textbf{EgoSchema} \citep{mangalam2023egoschema} is a long-form multiple-choice video question answering dataset derived from Ego4D \citep{grauman2022ego4d}. The task involves selecting the correct answer out of five options based on a three-minute-long video clip. This task is particularly interesting for evaluating long-context understanding, as it benefits from long ``temporal certificate'' lengths, \ie the minimum video duration a human needs to answer the question accurately.  The model is fine-tuned on a mixture of HowTo100M and Ego4D, and we ensure that there is no overlap between Ego4D training and EgoSchema examples.

\textbf{Next-QA} \citep{xiao2021nextqa} emphasizes testing causal and temporal reasoning with open- and close-ended (multiple-choice) QA tasks. Videos in this dataset have an average duration of 44 seconds but can be as long as 2 minutes. We use the close-ended version for both fine-tuning and inference. Since the training set is fairly small and in order to avoid over-fitting on this domain, we add and only tune low-rank adapters (LoRA;~\citealt{hulora}) at the self-attention and feed-forward blocks of every layer, which account for $\sim$12\% of model parameters. For fine-tuning on this multiple-choice QA dataset, we use the four incorrect answers to the given question as hard negatives in \cref{eq:con}.

\textbf{Perception Test} \citep{puatruaucean2023perception} is inspired by assessment in developmental psychology and features a collection of games or daily activities that evaluate a model's grasp of physics, reasoning, memory, and semantic extraction. Although videos in this dataset are short with an average duration of 30 seconds, accurate localization and recognition of actions and objects require a higher FPS rate (we use an FPS of $4$), resulting in sequences of hundreds of frames. We evaluate on the multiple-choice video question answering task by selecting one out of three possible answers, while training on Next-QA for zero-shot evaluation on this benchmark.

\begin{figure*}[t]
\begin{center}
\includegraphics[width=\textwidth]{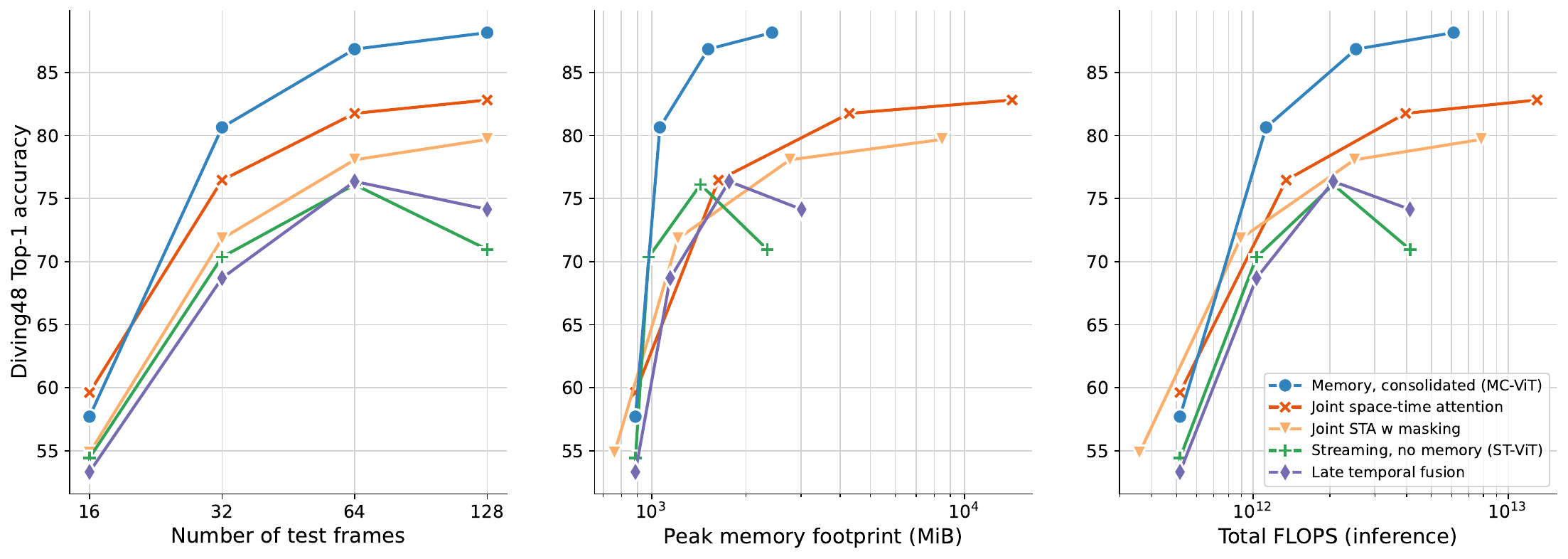}
\vspace{-2em}
\caption{\textbf{\ourmethod \ efficiently models long videos.} Fine-grained video understanding on Diving48 as a function of number of test frames (\textbf{left}), memory consumption (\textbf{middle}), and computational complexity (FLOPS, \textbf{right}), for joint space-time attention w/ and w/o masking (yellow and red respectively), memory-less streaming setting (green), the late temporal fusion baseline (purple) and our proposed method \ourmethod \ (blue). \ourmethod \ reaches the highest accuracy with 10$\x$ less memory and FLOPS than the joint space-time attention method. \vspace{-2em}}
\label{fig:inference}
\end{center}
\end{figure*}

\subsection{MC-ViT Effectively Learns from Long Videos}
\label{sec:exp-learning}

We start by assessing the ability of \ourmethod \ to model videos of increasing lengths. For this we fine-tune \ourmethod \ on videos with different number of frames (16, 32, 64, or 128) by varying the FPS rate. At inference time, we also apply the model to videos with 16 to 256 frames. \Cref{fig:learning} (left) shows that \ourmethod's performance improves with more, densely sampled frames at both training and inference time on Diving48 fine-grained action recognition. In particular, training with longer contexts allows \ourmethod \ to benefit from more frames at inference time, with the optimal inference-time video length being twice that of the train-time video length, demonstrating reasonable generalization of the consolidated cross-attention mechanism. 

In contrast, neither joint space-time attention (\Cref{fig:learning}, middle) nor a memory-less streaming ST-ViT architecture (\Cref{fig:learning}, right) effectively learn from long videos. While joint-space time attention benefits from training on more frames in terms of performance, its memory footprint prevents it from training or evaluating on the longest videos. ST-ViT on the other hand scales to more frames, but does not benefit from them, since it lacks the ability to reason over events that span multiple segments.

\subsection{MC-ViT Efficiently Models Long Videos}
\label{sec:exp-inference}

\textbf{Inference-time efficiency.} We evaluate the performance of joint space-time attention, ST-ViT, and MC-ViT in relation to their memory and computational complexity, by varying the number of frames at inference time (all models are trained with 64 frames). \ourmethod's memory consumption is bounded by the number of tokens within a segment, similar to memory-less ST-ViT, whereas that of joint space-time attention increases with video length (\Cref{fig:inference}, middle). Similarly, while the computational complexity of joint space-time attention is quadratic in the video length, it is linear for both ST-ViT and MC-ViT (\Cref{fig:inference}, right). 

In terms of performance, \Cref{fig:inference} demonstrates that \ourmethod \ remarkably outperforms joint space-time attention with a 10$\x$ smaller memory footprint (middle) and FLOPS (right). We additionally test other scalable baselines, such as 
applying 25\% input token masking to joint space-time attention~\cite{papalampidi2023simple}, and late temporal fusion~\cite{alayrac2022flamingo,yan2022video}, where we add a learnable module on top of ST-ViT for contextualizing information across segments (see Appendix~\ref{sec:app_comparison_methods}). Not only does \ourmethod \ display a better scaling behavior than these baselines (\Cref{fig:inference}, left), but it does so with robust improvements in memory footprint and computational complexity.

\begin{figure}[!thbp]
\begin{center}
\centerline{\includegraphics[width=\linewidth]{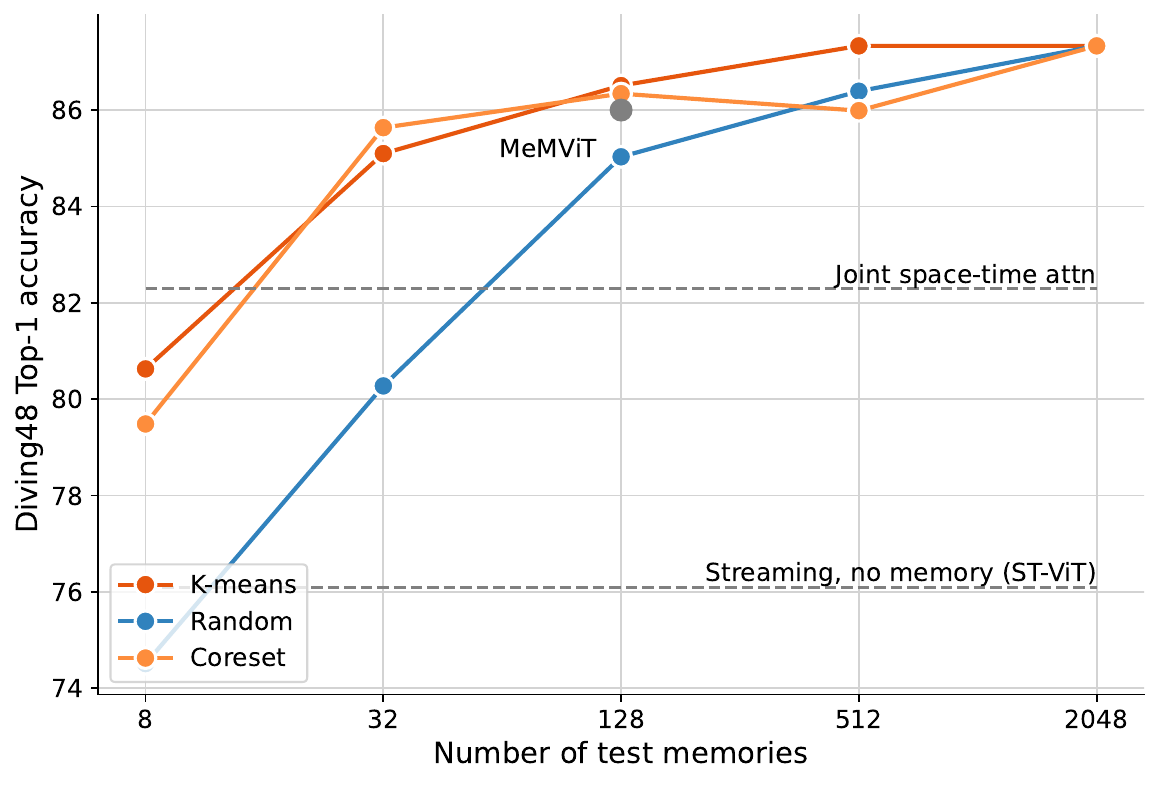}}
\vspace{-1em}
\caption{
\textbf{\ourmethod \ makes efficient use of finite-length context.} We show three MC-ViT instances and compare them to relevant baselines (dashed horizontal lines). K-means (red) and coreset (orange) surpass all methods at 16$\x$ compression rate with 128 memories per segment, demonstrating the efficiency of our approach. Surprisingly, even random memory selection (blue) achieves impressive performance on this task, outperforming all baselines at 4$\x$ compression rate with 512 memories, which further showcases efficiency and robustness of the MC-ViT  framework.
\vspace{-2em}
}
\label{fig:memory}
\end{center}
\end{figure}

\textbf{Training-time efficiency.} We additionally assess whether these inference-time gains in efficiency translate in similar gains when fine-tuning. Specifically, we vary the number of frames used for fine-tuning, and measure the resulting performance when evaluating the model with twice the number of frames (as suggested in \Cref{fig:learning}), as well as the training-time memory footprint and computational complexity. \Cref{fig:training_efficiency}, left, shows that \ourmethod\ quickly surpasses joint space-time attention and the streaming memory architecture when provided with longer videos for training. In particular, \ourmethod\  surpasses the performance of joint space-time attention with a 3.5$\x$ smaller memory footprint and 3.2$\x$ fewer FLOPS (\Cref{fig:training_efficiency}, middle and right).

\begin{table*}[t]
\caption{\textbf{MC-ViT with memory size decoupled from video length.} For MC-ViT-last, we store the last $N$ segments (of length $K \!=\! 512$) into memory. The equivalent for MC-ViT-global is to store randomly selected $N \!\times\! K$ tokens across segments. 
\vspace{0.1em}
}
\label{tab:fixed_memory}
\begin{center}
\begin{small}
\begin{tabular}{lcccccc}
& \multicolumn{6}{c}{$N$ / $N\!\times\!K$} \\
 \cmidrule(lr){2-7}
Method & 2 / 1024 & 3 / 1536 & 4 / 2048 & 5 / 2560 & 6 / 3072 &	7 / 3584 \\
\midrule
MC-ViT-B-last & 81.5 & 85.0 & 87.0 & \textbf{87.8} & \textbf{87.8} & \textbf{87.8} \\ 
MC-ViT-B-global & \textbf{84.8} & \textbf{86.8} & \textbf{87.5} & \textbf{87.8} & \textbf{87.8} & \textbf{87.8} \\ 
\end{tabular}
\end{small}
\end{center}
\vspace{-0.3em}
\end{table*}

\begin{table*}[t]
\caption{\textbf{Long video question answering, compared to public models.} Performance is calculated as percentage correct on multiple-choice video question answering on EgoSchema, Perception Test and Next-QA. By scaling to significantly longer videos, \ourmethod \ outperforms models that benefit from an order of magnitude more parameters. We highlight the \textbf{best} and \underline{second-best} methods per dataset. 128+ stands for 128 or more frames, where we use 128 frames for EgoSchema and Next-QA and 256 frames for Perception Test.}
\label{tab:public_vqa}
\begin{center}
\begin{small}
\begin{tabular}{lcccccc}
Method & Params & Frames & \multicolumn{2}{c}{EgoSchema} & Perception Test & Next-QA \\
\cmidrule(lr){4-5}
 &  &  &  Subset & Full &  \\
\midrule

\vspace{-0.75em} \\
CoVGT~\cite{xiao2023contrastive} & 149M & 32 & -- & -- & -- & 60.0 \\
SeViT\textsubscript{FiD} \cite{kim2023semi} & 215M & 10 & -- & -- & -- & 60.6 \\
HiTeA~\cite{ye2023hitea} & 297M & 16 & -- & -- & -- & 63.1 \\ 
InternVideo~\cite{wang2022internvideo} & 478M & 90 & -- & 32.1 & -- & 63.2 \\
ImageViT~\cite{papalampidi2023simple} & 1B & 16 & 40.8 & 30.9 & 39.1 & -- \\ 
ShortViViT~\cite{papalampidi2023simple} & 1B & 16 & 47.9 & 31.0 & 41.9 & -- \\
Flamingo \cite{alayrac2022flamingo} & 3B & 32 & -- & -- & 43.6 & -- \\ 
SeViLA Localizer + ShortViViT~\cite{papalampidi2023simple} & 5B & 32 & 49.6 & 31.3 & -- & -- \\
LongViViT~\cite{papalampidi2023simple} & 1B & 256 & 56.8 & 33.3 & 45.7 & -- \\ 
SeViLA~\cite{yu2023self} & 4B & 32 & 25.7 & 22.7 & 46.2 & \textbf{73.8} \\ 

\vspace{-0.75em} \\

\rowcolor{ourbg} \ourmethod-B & 203M & \hspace{0.5em}128+ & \underline{61.2} & \underline{42.3} & \underline{47.0} & 60.6 \\
\rowcolor{ourbg} \ourmethod-L & 424M & \hspace{0.5em}128+ & \textbf{62.6} & \textbf{44.4} & \textbf{48.1} & \underline{65.0} \\
\end{tabular}
\end{small}
\end{center}
\vspace{-1em}
\end{table*}

\begin{table}[t]
\caption{\textbf{Fine-grained action classification on Diving48.} Prior methods use 3$\x$ more spatial crops at inference time (SC) and/or bounding box information (BB), which MC-ViT does not require. \vspace{-1.0em}
}
\label{tab:diving}
\begin{center}
\begin{small}
\begin{tabular}{lccc}
Method & Params & Extra & Top-1 \\
\midrule

TimeS-L~\cite{bertasius2021space} & 121M & SC & 81.0 \\ 
VideoSwin-B~\cite{liu2022video} & 88M & SC & 81.9 \\ 
BEVT~\cite{wang2022bevt} & 88M & SC & 86.7 \\ 
SIFAR-B-14~\cite{fan2021can} & 87M & SC & 87.3 \\ 
ORViT~\cite{herzig2022object} & 160M & SC+BB & 88.0 \\ 
AIM ViT-B~\cite{yang2023aim} & 97M & SC & 88.9 \\ 
AIM ViT-L~\cite{yang2023aim} & 341M & SC & 90.6 \\ 
\vspace{-0.75em} \\
\rowcolor{ourbg} \ourmethod-B & 99M & \xmark & \textbf{89.7} \\
\rowcolor{ourbg} \ourmethod-L & 313M & \xmark & \textbf{91.0} \\
\end{tabular}
\end{small}
\end{center}
\vspace{1.0em}
\end{table}

\subsection{Memory Consolidation Makes Efficient Use of a Finite Context Window}
\label{sec:exp-memory}
\textbf{Consolidation mechanisms.} We now analyze the computational efficiency and expressiveness of \ourmethod's consolidation methods. We compare our methods to three baselines: (1) joint space-time attention, (2) ST-ViT, and (3) MeMViT \citep{wu2022memvit}. Notably, MeMViT employs a parametric approach to memory compression, requiring a convolutional module to be trained alongside the network, where we experimented with different convolutional kernel sizes and found the kernel size $4 \!\times\! 2 \!\times\! 2$ (equivalent to MC-ViT's $K \!=\! 128$) to perform the best (see Appendix~\ref{sec:app_comparison_methods} for details). \Cref{fig:memory} illustrates the performance of these methods on Diving48 as a function of the number of memories $K$ per segment. Given $K\!=\!128$ memories obtained through k-means consolidation (i.e. a 16$\times$ compression compared to MA-ViT; red curve), \ourmethod-KM\ outperforms all baselines. Remarkably, even random selection of $K\!=\!128$ memories (with \ourmethod-R) is sufficient to surpass ViViT and ST-ViT. Finally, consolidating past-activations with \ourmethod-CS\ (coreset, orange curve) performs similarly to \ourmethod-KM, highlighting the robustness of \ourmethod \ to the particular choice of memory consolidation algorithm. K-means consolidation is used as the default method given its greater computational efficiency
and slightly higher performance for larger sets of memories.

\textbf{Finite memory size.} Further, to adapt MC-ViT to very long videos and decouple memory growth from video length, we compare two simple modifications to MC-ViT in Table \ref{tab:fixed_memory} on the Diving-48 dataset: (1) keep all tokens from the last $N$ segments of length $K$ (MC-ViT-last) or (2) randomly select a fixed number of $N \!\times\! K$ tokens across all past segments (MC-ViT-global). The results show that 5 segments (or 2560 tokens) should be saved in memory with no loss in performance. Further, we see that selecting $N \!\times\! K$ random tokens across segments (with $K\!=\!512$ for all experiments) is more beneficial than keeping the last $N \!\times\! K$ tokens, which shows that relevant information for the current segment is contained throughout the whole video and not just in the last several segments. This also emphasizes the importance of redundancy reduction in the consolidation scheme, as uniformly sampled memories from the entire video are less redundant than all tokens from the most recent segments.

\begin{table*}[t]
\caption{\textbf{Long video question answering on EgoSchema and Perception Test, compared to large-scale proprietary models.} Performance is evaluated on the original (``raw'') dataset, as well as on the ``visual'' subset of questions that cannot be answered by a blind language model and on Perception Test for the validation set. For each model, we compute the performance of a ``blind'' variant  on EgoSchema that only has access to question-answer pairs. The performance of the blind model is subtracted from that of the full model to compute ``visual'' performance. We \underline{underline} the top 2 performing models for each benchmark and subset.
}
\label{tab:egoschema_proprietary}
\begin{center}
\begin{small}
\begin{tabular}{lcccccc}
 & \multicolumn{2}{c}{EgoSchema Raw} & \multicolumn{2}{c}{EgoSchema Visual} & \multirow{2}{*}{\shortstack{Perception \\ Test  Raw}} & \multirow{2}{*}{\shortstack{Perception \\ Test Visual}} \\
 \cmidrule(lr){2-3} \cmidrule(lr){4-5}
 
Method & Subset & Full & Subset & Full \\
\midrule

Random chance          & 20.0 & 20.0 & -- & -- & 33.3 & -- \\ 
\vspace{-0.2em} \\
\textcolor{darkgray}{Bard only (blind)}           & \textcolor{darkgray}{27.0} & \textcolor{darkgray}{33.2} & \hspace{0.5em}\textcolor{darkgray}{0.0} & \hspace{0.5em}\textcolor{darkgray}{0.0} & \textcolor{darkgray}{36.8} & \hspace{0.5em}\textcolor{darkgray}{0.0} \\ 
Bard + ImageViT \citep{papalampidi2023simple}
& 35.0 & 35.0 & \hspace{0.5em}8.0 & \hspace{0.5em}1.8 & 37.8 & \hspace{0.5em}1.0 \\
Bard + ShortViViT \citep{papalampidi2023simple}  & 42.0 & 36.2 & 15.0 & \hspace{0.5em}3.0 & 38.8 & \hspace{0.5em}2.0 \\
Bard + PALI \citep{papalampidi2023simple}  & 44.8 & 39.2 & 17.8 & \hspace{0.5em}6.0 & 42.4 & \hspace{0.5em}5.6 \\ 
\vspace{-0.35em} \\

\textcolor{darkgray}{GPT-4 Turbo (blind)}       & \textcolor{darkgray}{31.0} & \textcolor{darkgray}{30.8} & \hspace{0.5em}\textcolor{darkgray}{0.0}  & \hspace{0.5em}\textcolor{darkgray}{0.0} & \textcolor{darkgray}{--} & \textcolor{darkgray}{--} \\
GPT-4V & \underline{63.5} & \underline{55.6} & 32.5 & \underline{24.8} & -- & -- \\
Gemini Ultra~\cite{team2023gemini} & -- &  -- &  -- &  -- & \underline{54.7} & -- \\ 
\vspace{-0.35em} \\
\rowcolor{ourbg} \textcolor{darkgray}{\ourmethod-B (blind)} & \textcolor{darkgray}{18.2} & \textcolor{darkgray}{23.4} & \hspace{0.5em}\textcolor{darkgray}{0.0}  & \hspace{0.5em}\textcolor{darkgray}{0.0}  & \textcolor{darkgray}{37.6} & \hspace{0.5em}\textcolor{darkgray}{0.0}  \\
\rowcolor{ourbg} \ourmethod-B & 61.2 & 42.3 & \underline{43.0} & 18.9 & 47.1 & \hspace{0.5em}\underline{9.5}  \\
\rowcolor{ourbg} \textcolor{darkgray}{\ourmethod-L (blind)} & \textcolor{darkgray}{15.0} & \textcolor{darkgray}{22.7} & \hspace{0.5em}\textcolor{darkgray}{0.0}  & \hspace{0.5em}\textcolor{darkgray}{0.0}  & \textcolor{darkgray}{35.1} & \hspace{0.5em}\textcolor{darkgray}{0.0}  \\
\rowcolor{ourbg} \ourmethod-L & \underline{62.6} & \underline{44.0} & \underline{47.6} & \underline{21.3} &
\underline{47.6} & \underline{12.5} \\


\end{tabular}
\end{small}
\end{center}
\vskip -0.1in
\end{table*}

\subsection{MC-ViT Achieves State-of-the-Art Long-Context Video Understanding}
\label{sec:exp-sota}

\noindent \textbf{Fine-grained action recognition.} In \Cref{tab:diving}, we compare \ourmethod \ to prior methods on Diving48, and find that it delivers state-of-the-art results. Unlike previous methods that require object tracking models \citep{herzig2022object} or additional modeling components, \ourmethod \ achieves strong performance by simply re-purposing a general transformer architecture for long-context modeling: while previous methods are limited to 32 frames of video, the efficient scaling properties of MC-ViT allow it to process 128 frames. Further, MC-ViT does not require multiple spatial crops at inference time to achieve state-of-the-art results.

\noindent \textbf{Long video question answering.}
We compare MC-ViT to prior methods on long video question answering in Table~\ref{tab:public_vqa}. We find that our approach outperforms prior works that use up to 10$\x$ more parameters. Most notably, even our smaller model version (MC-ViT-B, with 200M parameters in total) is able to achieve a $10\%$ improvement on EgoSchema in comparison to much larger models (up to 5B parameters). This demonstrates the importance of processing more frames, which our straightforward memory consolidation method enables, as well as the effectiveness of fine-tuning \ourmethod \ from standard pretrained video encoders.

It is particularly notable that \ourmethod \ is competitive with models such as Flamingo~\citep{alayrac2022flamingo} and SeViLA~\citep{yu2023self}, which boast billion-scale LLM decoders. Such methods benefit from the language bias in VQA---which allows for some questions to be trivially answered without any visual input---and extensive textual training data. 
While \ourmethod \ surpasses these models on EgoSchema and Perception Test, SeViLa maintains stronger performance on Next-QA. 
We hypothesize that this benchmark is not challenging enough for long video understanding and relies heavily on language-only reasoning, since \citet{yu2023self} achieve their results while using a single input frame. Thus, frame-level models with strong decoders, such as SeViLA, may be sufficient for benchmarks requiring language-only reasoning and localization (Next-QA, Perception Test), but fail to capture a summary representation of the entire video (EgoSchema). In contrast, our method, despite lacking large language decoders, performs competitively across the board, demonstrating strong localization and long-context modeling capabilities. Finally, \ourmethod~requires minimal architectural changes and training overhead for adapting to long-context understanding, in contrast to modular methods~\citep[e.g.,][]{yu2023self} which involve multiple modules and complex training regimes.

\noindent \textbf{MC-ViT vs. large-scale proprietary models.} 
Additionally, in \Cref{tab:egoschema_proprietary} we compare our method to large-scale proprietary models such as GPT-4V \citep{achiam2023gpt4}, Gemini~\citep{team2023gemini} and Bard\footnote{Release of September 2023.} + PALI \citep{Bard2023,chen2022pali}. While their exact implementation details are not publicly available, these models are thought to contain hundreds of billions to trillions of parameters, \ie 1000$\x$ more than \ourmethod. It is also important to note that these proprietary models are trained on massive amounts of data from the internet, resulting in potential data contamination, which we proactively avoid in our training pipeline. %

We showcase the performance of all models in the ``raw'' columns. Additionally, we also present (in grey font) the blind model performance where the model is trained only on question-answer pairs without the visual modality. In order to isolate the visual perception capabilities from the natural language reasoning of the model, we add the ``visual'' column in \cref{tab:egoschema_proprietary} which is computed as the difference between ``raw'' and ``blind''.
Examining the visual-only capabilities, we conclude that our small-scale model is competitive against the large proprietary ones and even surpasses GPT-4V performance on the subset of EgoSchema.

Despite using a fraction of the parameters and training data, our method remains competitive and, in some cases, outperforms these models. In particular, MC-ViT achieves 5\% improvements on EgoSchema and Perception Test against the sophisticated Bard + PALI modular system used for information aggregation and frame captioning, respectively. 

\section{Discussion}

In this work, we introduced the Memory-Consolidated Vision Transformer (MC-ViT), which efficiently models long-range dependencies in videos by consolidating past activations into a compact memory bank. 
MC-ViT achieves state-of-the-art performance on multiple long video benchmarks by repurposing existing video architectures without the need for specialized architectures and training regimes. Our small-scale model
outperforms approaches that benefit from orders of magnitude more parameters, and is even competitive with large-scale proprietary systems such as GPT-4V and Bard, demonstrating the importance of strong compressed video representations. As an extension, these representations could be fed into large language models to augment their long-range temporal reasoning capabilities.

We showcased the effectiveness of non-parametric memory consolidation techniques as a simple means of extending long video contexts, and future work could straightforwardly build on \ourmethod \ by exploring alternative consolidation strategies.
For instance, incorporating insights from cognitive models of memory, such as the role of episodic and semantic memory systems, as well as theories of efficient coding \cite{barlow1961possible}, could inspire new consolidation techniques. Furthermore, the concept of memory consolidation could be applied to other domains involving sequential data, such as natural language and audio processing, laying the foundation for personalized assistant technologies that jointly reason over multiple modalities. 

\section*{Impact Statement}
By adapting standard video architectures to the long-context setup, this work could potentially equip general-purpose assistant models with the ability to efficiently process long videos. These models will likely suffer from similar biases and potential harms associated with visual language models and large language models more generally.

Further, since this work focuses on efficient processing of long sequences without sacrificing performance, the corresponding methods and findings from this work could potentially be applied to other domains, such as NLP or audio, allowing for faster processing of large amounts of data and thus making long-context model training more readily available for widespread use.

\section*{Acknowledgements}
We thank Andrew Zisserman, João Carreira, Carl Allen, and Nikhil Parthasarathy for their thoughtful feedback, Relja Arandjelovi\'c for fruitful discussions at the inception of this project, and Oliver Vikbladh, Eleanor Spens, and Neil Burgess for sharing their insights into memory consolidation in the human mind. 

\bibliography{main}

\begin{thebibliography}{75}
\providecommand{\natexlab}[1]{#1}
\providecommand{\url}[1]{\texttt{#1}}
\expandafter\ifx\csname urlstyle\endcsname\relax
  \providecommand{\doi}[1]{doi: #1}\else
  \providecommand{\doi}{doi: \begingroup \urlstyle{rm}\Url}\fi

\bibitem[Achiam et~al.(2023)Achiam, Adler, Agarwal, Ahmad, Akkaya, Aleman,
  Almeida, Altenschmidt, Altman, Anadkat, et~al.]{achiam2023gpt4}
Achiam, J., Adler, S., Agarwal, S., Ahmad, L., Akkaya, I., Aleman, F.~L.,
  Almeida, D., Altenschmidt, J., Altman, S., Anadkat, S., et~al.
\newblock {GPT-4 technical report}.
\newblock \emph{arXiv preprint arXiv:2303.08774}, 2023.

\bibitem[Agarwal et~al.(2005)Agarwal, Har-Peled, Varadarajan,
  et~al.]{agarwal2005geometric}
Agarwal, P.~K., Har-Peled, S., Varadarajan, K.~R., et~al.
\newblock Geometric approximation via coresets.
\newblock \emph{Combinatorial and Computational Geometry}, 52\penalty0
  (1):\penalty0 1--30, 2005.

\bibitem[Alayrac et~al.(2022)Alayrac, Donahue, Luc, Miech, Barr, Hasson, Lenc,
  Mensch, Millican, Reynolds, et~al.]{alayrac2022flamingo}
Alayrac, J.-B., Donahue, J., Luc, P., Miech, A., Barr, I., Hasson, Y., Lenc,
  K., Mensch, A., Millican, K., Reynolds, M., et~al.
\newblock Flamingo: A visual language model for few-shot learning.
\newblock \emph{Advances in Neural Information Processing Systems}, 2022.

\bibitem[Anil et~al.(2023)Anil, Borgeaud, Wu, Alayrac, Yu, Soricut, Schalkwyk,
  Dai, Hauth, et~al.]{team2023gemini}
Anil, R., Borgeaud, S., Wu, Y., Alayrac, J.-B., Yu, J., Soricut, R., Schalkwyk,
  J., Dai, A.~M., Hauth, A., et~al.
\newblock {Gemini: A family of highly capable multimodal models}.
\newblock \emph{arXiv preprint arXiv:2312.11805}, 2023.

\bibitem[Arnab et~al.(2021)Arnab, Dehghani, Heigold, Sun, Lu{\v{c}}i{\'c}, and
  Schmid]{arnab2021vivit}
Arnab, A., Dehghani, M., Heigold, G., Sun, C., Lu{\v{c}}i{\'c}, M., and Schmid,
  C.
\newblock {ViViT: A video vision transformer}.
\newblock In \emph{Proceedings of the IEEE/CVF International Conference on
  Computer Vision}, 2021.

\bibitem[Ba et~al.(2016)Ba, Kiros, and Hinton]{ba2016layernorm}
Ba, J.~L., Kiros, J.~R., and Hinton, G.~E.
\newblock Layer normalization.
\newblock \emph{arXiv preprint arXiv:1607.06450}, 2016.

\bibitem[Barlow(1961)]{barlow1961possible}
Barlow, H.~B.
\newblock Possible principles underlying the transformation of sensory
  messages.
\newblock \emph{Sensory communication}, 1\penalty0 (01):\penalty0 217--233,
  1961.

\bibitem[Bartlett(1932)]{bartlett1932remembering}
Bartlett, F.~C.
\newblock \emph{Remembering: A study in experimental and social psychology}.
\newblock Cambridge University Press, 1932.

\bibitem[Beltagy et~al.(2020)Beltagy, Peters, and Cohan]{Beltagy2020Longformer}
Beltagy, I., Peters, M.~E., and Cohan, A.
\newblock Longformer: The long-document transformer.
\newblock \emph{arXiv:2004.05150}, 2020.

\bibitem[Bertasius et~al.(2021)Bertasius, Wang, and
  Torresani]{bertasius2021space}
Bertasius, G., Wang, H., and Torresani, L.
\newblock Is space-time attention all you need for video understanding?
\newblock In \emph{International Conference on Machine Learning}, 2021.

\bibitem[Bolya et~al.(2022)Bolya, Fu, Dai, Zhang, and Hoffman]{bolya2022hydra}
Bolya, D., Fu, C.-Y., Dai, X., Zhang, P., and Hoffman, J.
\newblock Hydra attention: Efficient attention with many heads.
\newblock In \emph{European Conference on Computer Vision}, 2022.

\bibitem[Borgeaud et~al.(2022)Borgeaud, Mensch, Hoffmann, Cai, Rutherford,
  Millican, Van Den~Driessche, Lespiau, Damoc, Clark,
  et~al.]{borgeaud2022improving}
Borgeaud, S., Mensch, A., Hoffmann, J., Cai, T., Rutherford, E., Millican, K.,
  Van Den~Driessche, G.~B., Lespiau, J.-B., Damoc, B., Clark, A., et~al.
\newblock Improving language models by retrieving from trillions of tokens.
\newblock In \emph{International Conference on Machine Learning}, 2022.

\bibitem[Carreira \& Zisserman(2017)Carreira and Zisserman]{carreira2017quo}
Carreira, J. and Zisserman, A.
\newblock Quo vadis, action recognition? a new model and the kinetics dataset.
\newblock In \emph{Proceedings of the IEEE Conference on Computer Vision and
  Pattern Recognition}, 2017.

\bibitem[Chen et~al.(2023)Chen, Wang, Changpinyo, Piergiovanni, Padlewski,
  Salz, Goodman, Grycner, Mustafa, Beyer, et~al.]{chen2022pali}
Chen, X., Wang, X., Changpinyo, S., Piergiovanni, A., Padlewski, P., Salz, D.,
  Goodman, S., Grycner, A., Mustafa, B., Beyer, L., et~al.
\newblock {PaLI: A jointly-scaled multilingual language-image model}.
\newblock In \emph{International Conference on Learning Representations}, 2023.

\bibitem[Dai et~al.(2017)Dai, Yang, Yang, Cohen, and
  Salakhutdinov]{dai2017good}
Dai, Z., Yang, Z., Yang, F., Cohen, W.~W., and Salakhutdinov, R.~R.
\newblock Good semi-supervised learning that requires a bad gan.
\newblock In \emph{Advances in Neural Information Processing Systems}, 2017.

\bibitem[Dai et~al.(2019)Dai, Yang, Yang, Carbonell, Le, and
  Salakhutdinov]{dai2019transformer}
Dai, Z., Yang, Z., Yang, Y., Carbonell, J., Le, Q.~V., and Salakhutdinov, R.
\newblock {Transformer-XL: Attentive language models beyond a fixed-length
  context}.
\newblock In \emph{Proceedings of the Association for Computational
  Linguistics}, 2019.

\bibitem[Dao et~al.(2022)Dao, Fu, Ermon, Rudra, and
  R{\'e}]{dao2022flashattention}
Dao, T., Fu, D., Ermon, S., Rudra, A., and R{\'e}, C.
\newblock {FlashAttention: Fast and memory-efficient exact attention with
  IO-awareness}.
\newblock \emph{Advances in Neural Information Processing Systems}, 2022.

\bibitem[Dong et~al.(2022)Dong, Bao, Chen, Zhang, Yu, Yuan, Chen, and
  Guo]{dong2022cswin}
Dong, X., Bao, J., Chen, D., Zhang, W., Yu, N., Yuan, L., Chen, D., and Guo, B.
\newblock {CSWin transformer: A general vision transformer backbone with
  cross-shaped windows}.
\newblock In \emph{Proceedings of the IEEE/CVF Conference on Computer Vision
  and Pattern Recognition}, 2022.

\bibitem[Dosovitskiy et~al.(2021)Dosovitskiy, Beyer, Kolesnikov, Weissenborn,
  Zhai, Unterthiner, Dehghani, Minderer, Heigold, Gelly,
  et~al.]{dosovitskiy2020image}
Dosovitskiy, A., Beyer, L., Kolesnikov, A., Weissenborn, D., Zhai, X.,
  Unterthiner, T., Dehghani, M., Minderer, M., Heigold, G., Gelly, S., et~al.
\newblock An image is worth 16x16 words: Transformers for image recognition at
  scale.
\newblock In \emph{International Conference on Learning Representations}, 2021.

\bibitem[Fan et~al.(2021)Fan, Panda, et~al.]{fan2021can}
Fan, Q., Panda, R., et~al.
\newblock Can an image classifier suffice for action recognition?
\newblock In \emph{International Conference on Learning Representations}, 2021.

\bibitem[Feichtenhofer et~al.(2019)Feichtenhofer, Fan, Malik, and
  He]{feichtenhofer2019slowfast}
Feichtenhofer, C., Fan, H., Malik, J., and He, K.
\newblock {SlowFast networks for video recognition}.
\newblock In \emph{Proceedings of the IEEE/CVF International Conference on
  Computer Vision}, 2019.

\bibitem[{Google AI}(2023)]{Bard2023}
{Google AI}.
\newblock Bard [large language model], 2023.

\bibitem[Grauman et~al.(2022)Grauman, Westbury, Byrne, Chavis, Furnari,
  Girdhar, Hamburger, Jiang, Liu, Liu, et~al.]{grauman2022ego4d}
Grauman, K., Westbury, A., Byrne, E., Chavis, Z., Furnari, A., Girdhar, R.,
  Hamburger, J., Jiang, H., Liu, M., Liu, X., et~al.
\newblock {Ego4D: Around the world in 3,000 hours of egocentric video}.
\newblock In \emph{Proceedings of the IEEE/CVF Conference on Computer Vision
  and Pattern Recognition}, 2022.

\bibitem[Graves et~al.(2014)Graves, Wayne, and Danihelka]{graves2014neural}
Graves, A., Wayne, G., and Danihelka, I.
\newblock {Neural Turing machines}.
\newblock \emph{arXiv preprint arXiv:1410.5401}, 2014.

\bibitem[Han et~al.(2020)Han, Xie, and Zisserman]{han2020memory}
Han, T., Xie, W., and Zisserman, A.
\newblock Memory-augmented dense predictive coding for video representation
  learning.
\newblock In \emph{Proceedings of the European Conference on Computer Vision},
  2020.

\bibitem[Herzig et~al.(2022)Herzig, Ben-Avraham, Mangalam, Bar, Chechik,
  Rohrbach, Darrell, and Globerson]{herzig2022object}
Herzig, R., Ben-Avraham, E., Mangalam, K., Bar, A., Chechik, G., Rohrbach, A.,
  Darrell, T., and Globerson, A.
\newblock Object-region video transformers.
\newblock In \emph{Proceedings of the IEEE/CVF Conference on Computer Vision
  and Pattern Recognition}, 2022.

\bibitem[Hu et~al.(2021)Hu, Wallis, Allen-Zhu, Li, Wang, Wang, Chen,
  et~al.]{hulora}
Hu, E.~J., Wallis, P., Allen-Zhu, Z., Li, Y., Wang, S., Wang, L., Chen, W.,
  et~al.
\newblock Lora: Low-rank adaptation of large language models.
\newblock In \emph{International Conference on Learning Representations}, 2021.

\bibitem[Jaegle et~al.(2022)Jaegle, Borgeaud, Alayrac, Doersch, Ionescu, Ding,
  Koppula, Zoran, Brock, Shelhamer, et~al.]{jaegle2021perceiverio}
Jaegle, A., Borgeaud, S., Alayrac, J.-B., Doersch, C., Ionescu, C., Ding, D.,
  Koppula, S., Zoran, D., Brock, A., Shelhamer, E., et~al.
\newblock {PerceiverIO: A general architecture for structured inputs \&
  outputs}.
\newblock In \emph{International Conference on Learning Representations}, 2022.

\bibitem[Jia et~al.(2022)Jia, Tang, Chen, Cardie, Belongie, Hariharan, and
  Lim]{jia2022visual}
Jia, M., Tang, L., Chen, B.-C., Cardie, C., Belongie, S., Hariharan, B., and
  Lim, S.-N.
\newblock Visual prompt tuning.
\newblock In \emph{Proceedings of the European Conference on Computer Vision},
  2022.

\bibitem[Kim et~al.(2023)Kim, Kim, Lee, and Seo]{kim2023semi}
Kim, S., Kim, J.-H., Lee, J., and Seo, M.
\newblock Semi-parametric video-grounded text generation.
\newblock \emph{arXiv preprint arXiv:2301.11507}, 2023.

\bibitem[Lai et~al.(2020)Lai, Lu, and Xie]{lai2020mast}
Lai, Z., Lu, E., and Xie, W.
\newblock {MAST: A memory-augmented self-supervised tracker}.
\newblock In \emph{Proceedings of the IEEE/CVF Conference on Computer Vision
  and Pattern Recognition}, 2020.

\bibitem[Li et~al.(2023)Li, He, Wang, Li, Wang, Luo, Wang, Wang, and
  Qiao]{li2023videochat}
Li, K., He, Y., Wang, Y., Li, Y., Wang, W., Luo, P., Wang, Y., Wang, L., and
  Qiao, Y.
\newblock {VideoChat}: Chat-centric video understanding.
\newblock \emph{arXiv preprint arXiv:2305.06355}, 2023.

\bibitem[Li et~al.(2018)Li, Li, and Vasconcelos]{li2018diving48}
Li, Y., Li, Y., and Vasconcelos, N.
\newblock {RESOUND: Towards action recognition without representation bias}.
\newblock In \emph{Proceedings of the European Conference on Computer Vision},
  2018.

\bibitem[Li et~al.(2022{\natexlab{a}})Li, Mao, Girshick, and
  He]{li2022exploring}
Li, Y., Mao, H., Girshick, R., and He, K.
\newblock Exploring plain vision transformer backbones for object detection.
\newblock In \emph{Proceedings of the European Conference on Computer Vision},
  2022{\natexlab{a}}.

\bibitem[Li et~al.(2022{\natexlab{b}})Li, Wu, Fan, Mangalam, Xiong, Malik, and
  Feichtenhofer]{li2022mvitv2}
Li, Y., Wu, C.-Y., Fan, H., Mangalam, K., Xiong, B., Malik, J., and
  Feichtenhofer, C.
\newblock {MViTv2: Improved multiscale vision transformers for classification
  and detection}.
\newblock In \emph{Proceedings of the IEEE/CVF Conference on Computer Vision
  and Pattern Recognition}, 2022{\natexlab{b}}.

\bibitem[Lin et~al.(2023)Lin, Ahmed, Li, Lin, Azarnasab, Yang, Wang, Liang,
  Liu, Lu, et~al.]{lin2023mmvid}
Lin, K., Ahmed, F., Li, L., Lin, C.-C., Azarnasab, E., Yang, Z., Wang, J.,
  Liang, L., Liu, Z., Lu, Y., et~al.
\newblock {MM-VID: Advancing video understanding with GPT-4V(ision)}.
\newblock \emph{arXiv preprint arXiv:2310.19773}, 2023.

\bibitem[Liu et~al.(2022)Liu, Ning, Cao, Wei, Zhang, Lin, and Hu]{liu2022video}
Liu, Z., Ning, J., Cao, Y., Wei, Y., Zhang, Z., Lin, S., and Hu, H.
\newblock {Video Swin transformer}.
\newblock In \emph{Proceedings of the IEEE/CVF Conference on Computer Vision
  and Pattern Recognition}, 2022.

\bibitem[Mangalam et~al.(2023)Mangalam, Akshulakov, and
  Malik]{mangalam2023egoschema}
Mangalam, K., Akshulakov, R., and Malik, J.
\newblock {EgoSchema: A diagnostic benchmark for very long-form video language
  understanding}.
\newblock In \emph{Advances in Neural Information Processing Systems}, 2023.

\bibitem[Marr(1971)]{marr1971simple}
Marr, D.
\newblock \emph{Simple memory: a theory for archicortex}.
\newblock Philosophical Transactions of the Royal Society of London, 1971.

\bibitem[Martins et~al.(2022)Martins, Marinho, and Martins]{martins2022former}
Martins, P.~H., Marinho, Z., and Martins, A.~F.
\newblock $\infty$-former: Infinite memory transformer.
\newblock In \emph{Proceedings of the Association for Computational
  Linguistics}, 2022.

\bibitem[Miech et~al.(2019)Miech, Zhukov, Alayrac, Tapaswi, Laptev, and
  Sivic]{miech19howto100m}
Miech, A., Zhukov, D., Alayrac, J.-B., Tapaswi, M., Laptev, I., and Sivic, J.
\newblock Howto100{M}: Learning a text-video embedding by watching hundred
  million narrated video clips.
\newblock In \emph{Proceedings of the IEEE/CVF International Conference on
  Computer Vision}, 2019.

\bibitem[Oh et~al.(2019)Oh, Lee, Xu, and Kim]{oh2019video}
Oh, S.~W., Lee, J.-Y., Xu, N., and Kim, S.~J.
\newblock Video object segmentation using space-time memory networks.
\newblock In \emph{Proceedings of the IEEE/CVF International Conference on
  Computer Vision}, 2019.

\bibitem[Papalampidi et~al.(2024)Papalampidi, Koppula, Pathak, Chiu, Heyward,
  Patraucean, Shen, Miech, Zisserman, and Nematzdeh]{papalampidi2023simple}
Papalampidi, P., Koppula, S., Pathak, S., Chiu, J., Heyward, J., Patraucean,
  V., Shen, J., Miech, A., Zisserman, A., and Nematzdeh, A.
\newblock A simple recipe for contrastively pre-training video-first encoders
  beyond 16 frames.
\newblock In \emph{Proceedings of the IEEE/CVF Conference on Computer Vision
  and Pattern Recognition}, 2024.

\bibitem[P{\u{a}}tr{\u{a}}ucean et~al.(2023)P{\u{a}}tr{\u{a}}ucean, Smaira,
  Gupta, Continente, Markeeva, Banarse, Koppula, Heyward, Malinowski, Yang,
  et~al.]{puatruaucean2023perception}
P{\u{a}}tr{\u{a}}ucean, V., Smaira, L., Gupta, A., Continente, A.~R., Markeeva,
  L., Banarse, D., Koppula, S., Heyward, J., Malinowski, M., Yang, Y., et~al.
\newblock Perception {T}est: A diagnostic benchmark for multimodal video
  models.
\newblock In \emph{Advances in Neural Information Processing Systems}, 2023.

\bibitem[Piergiovanni et~al.(2023)Piergiovanni, Kuo, and
  Angelova]{piergiovanni2023rethinking}
Piergiovanni, A., Kuo, W., and Angelova, A.
\newblock {Rethinking video ViTs: Sparse video tubes for joint image and video
  learning}.
\newblock In \emph{Proceedings of the IEEE/CVF Conference on Computer Vision
  and Pattern Recognition}, 2023.

\bibitem[Piergiovanni et~al.(2024)Piergiovanni, Nobel, Kim, Ryoo, Gomes, and
  Angelova]{piergiovanni2023mirasol3b}
Piergiovanni, A., Nobel, I., Kim, D., Ryoo, M.~S., Gomes, V., and Angelova, A.
\newblock Mirasol3{B}: A multimodal autoregressive model for time-aligned and
  contextual modalities.
\newblock \emph{Proceedings of the IEEE/CVF Conference on Computer Vision and
  Pattern Recognitio}, 2024.

\bibitem[Radford et~al.(2021)Radford, Kim, Hallacy, Ramesh, Goh, Agarwal,
  Sastry, Askell, Mishkin, Clark, et~al.]{radford2021learning}
Radford, A., Kim, J.~W., Hallacy, C., Ramesh, A., Goh, G., Agarwal, S., Sastry,
  G., Askell, A., Mishkin, P., Clark, J., et~al.
\newblock Learning transferable visual models from natural language
  supervision.
\newblock In \emph{International Conference on Machine Learning}, 2021.

\bibitem[Rae et~al.(2020)Rae, Potapenko, Jayakumar, and
  Lillicrap]{rae2019compressive}
Rae, J.~W., Potapenko, A., Jayakumar, S.~M., and Lillicrap, T.~P.
\newblock Compressive transformers for long-range sequence modelling.
\newblock \emph{International Conference on Learning Representations}, 2020.

\bibitem[Ryali et~al.(2023)Ryali, Hu, Bolya, Wei, Fan, Huang, Aggarwal,
  Chowdhury, Poursaeed, Hoffman, et~al.]{ryali2023hiera}
Ryali, C., Hu, Y.-T., Bolya, D., Wei, C., Fan, H., Huang, P.-Y., Aggarwal, V.,
  Chowdhury, A., Poursaeed, O., Hoffman, J., et~al.
\newblock Hiera: A hierarchical vision transformer without the
  bells-and-whistles.
\newblock In \emph{International Conference on Machine learning}, 2023.

\bibitem[Ryoo et~al.(2023)Ryoo, Gopalakrishnan, Kahatapitiya, Xiao, Rao, Stone,
  Lu, Ibarz, and Arnab]{ryoo2023token}
Ryoo, M.~S., Gopalakrishnan, K., Kahatapitiya, K., Xiao, T., Rao, K., Stone,
  A., Lu, Y., Ibarz, J., and Arnab, A.
\newblock Token {T}uring machines.
\newblock In \emph{Proceedings of the IEEE/CVF Conference on Computer Vision
  and Pattern Recognition}, 2023.

\bibitem[Song et~al.(2024)Song, Chai, Wang, Zhang, Zhou, Wu, Guo, Ye, Lu,
  Hwang, et~al.]{song2023moviechat}
Song, E., Chai, W., Wang, G., Zhang, Y., Zhou, H., Wu, F., Guo, X., Ye, T., Lu,
  Y., Hwang, J.-N., et~al.
\newblock {MovieChat: From dense token to sparse memory for long video
  understanding}.
\newblock In \emph{Proceedings of the IEEE/CVF Conference on Computer Vision
  and Pattern Recognition}, 2024.

\bibitem[Spens \& Burgess(2024)Spens and Burgess]{spens2024generative}
Spens, E. and Burgess, N.
\newblock A generative model of memory construction and consolidation.
\newblock \emph{Nature Human Behaviour}, pp.\  1--18, 2024.

\bibitem[Sun et~al.(2017)Sun, Shrivastava, Singh, and Gupta]{sun2017revisiting}
Sun, C., Shrivastava, A., Singh, S., and Gupta, A.
\newblock Revisiting unreasonable effectiveness of data in deep learning era.
\newblock In \emph{Proceedings of the IEEE International Conference on Computer
  Vision}, 2017.

\bibitem[Tulving(1985)]{tulving1985memory}
Tulving, E.
\newblock Memory and consciousness.
\newblock \emph{Canadian Psychology/Psychologie canadienne}, 26\penalty0
  (1):\penalty0 1, 1985.

\bibitem[Vaswani et~al.(2017)Vaswani, Shazeer, Parmar, Uszkoreit, Jones, Gomez,
  Kaiser, and Polosukhin]{vaswani2017attention}
Vaswani, A., Shazeer, N., Parmar, N., Uszkoreit, J., Jones, L., Gomez, A.~N.,
  Kaiser, {\L}., and Polosukhin, I.
\newblock Attention is all you need.
\newblock In \emph{Advances in Neural Information Processing Systems}, 2017.

\bibitem[Wang et~al.(2022{\natexlab{a}})Wang, Chen, Wu, Chen, Dai, Liu, Jiang,
  Zhou, and Yuan]{wang2022bevt}
Wang, R., Chen, D., Wu, Z., Chen, Y., Dai, X., Liu, M., Jiang, Y.-G., Zhou, L.,
  and Yuan, L.
\newblock {BEVT}: {BERT} pretraining of video transformers.
\newblock In \emph{Proceedings of the IEEE/CVF Conference on Computer Vision
  and Pattern Recognition}, 2022{\natexlab{a}}.

\bibitem[Wang et~al.(2023{\natexlab{a}})Wang, Zhao, Do, Agarwal, Lee, and
  Sun]{wang2023vamos}
Wang, S., Zhao, Q., Do, M.~Q., Agarwal, N., Lee, K., and Sun, C.
\newblock Vamos: Versatile action models for video understanding.
\newblock \emph{arXiv preprint arXiv:2311.13627}, 2023{\natexlab{a}}.

\bibitem[Wang et~al.(2021)Wang, Xie, Li, Fan, Song, Liang, Lu, Luo, and
  Shao]{wang2021pyramid}
Wang, W., Xie, E., Li, X., Fan, D.-P., Song, K., Liang, D., Lu, T., Luo, P.,
  and Shao, L.
\newblock Pyramid vision transformer: A versatile backbone for dense prediction
  without convolutions.
\newblock In \emph{Proceedings of the IEEE/CVF International Conference on
  Computer Vision}, 2021.

\bibitem[Wang et~al.(2023{\natexlab{b}})Wang, Dong, Cheng, Liu, Yan, Gao, and
  Wei]{wang2023augmenting}
Wang, W., Dong, L., Cheng, H., Liu, X., Yan, X., Gao, J., and Wei, F.
\newblock Augmenting language models with long-term memory.
\newblock \emph{Advances in Neural Information Processing Systems},
  2023{\natexlab{b}}.

\bibitem[Wang et~al.(2022{\natexlab{b}})Wang, Li, Li, He, Huang, Zhao, Zhang,
  Xu, Liu, Wang, et~al.]{wang2022internvideo}
Wang, Y., Li, K., Li, Y., He, Y., Huang, B., Zhao, Z., Zhang, H., Xu, J., Liu,
  Y., Wang, Z., et~al.
\newblock Intern{V}ideo: General video foundation models via generative and
  discriminative learning.
\newblock \emph{arXiv preprint arXiv:2212.03191}, 2022{\natexlab{b}}.

\bibitem[Wang et~al.(2022{\natexlab{c}})Wang, Li, Xu, Zhou, Lei, Lin, Wang,
  Yang, Zhu, Hoiem, et~al.]{wang2022language}
Wang, Z., Li, M., Xu, R., Zhou, L., Lei, J., Lin, X., Wang, S., Yang, Z., Zhu,
  C., Hoiem, D., et~al.
\newblock Language models with image descriptors are strong few-shot
  video-language learners.
\newblock \emph{Advances in Neural Information Processing Systems},
  2022{\natexlab{c}}.

\bibitem[Wu et~al.(2019)Wu, Feichtenhofer, Fan, He, Krahenbuhl, and
  Girshick]{wu2019long}
Wu, C.-Y., Feichtenhofer, C., Fan, H., He, K., Krahenbuhl, P., and Girshick, R.
\newblock Long-term feature banks for detailed video understanding.
\newblock In \emph{Proceedings of the IEEE/CVF Conference on Computer Vision
  and Pattern Recognition}, 2019.

\bibitem[Wu et~al.(2022)Wu, Li, Mangalam, Fan, Xiong, Malik, and
  Feichtenhofer]{wu2022memvit}
Wu, C.-Y., Li, Y., Mangalam, K., Fan, H., Xiong, B., Malik, J., and
  Feichtenhofer, C.
\newblock {MeMViT}: Memory-augmented multiscale vision transformer for
  efficient long-term video recognition.
\newblock In \emph{Proceedings of the IEEE/CVF Conference on Computer Vision
  and Pattern Recognition}, 2022.

\bibitem[Xiao et~al.(2021)Xiao, Shang, Yao, and Chua]{xiao2021nextqa}
Xiao, J., Shang, X., Yao, A., and Chua, T.-S.
\newblock {Next-QA}: Next phase of question-answering to explaining temporal
  actions.
\newblock In \emph{Proceedings of the IEEE/CVF Conference on Computer Vision
  and Pattern Recognition}, 2021.

\bibitem[Xiao et~al.(2023)Xiao, Zhou, Yao, Li, Hong, Yan, and
  Chua]{xiao2023contrastive}
Xiao, J., Zhou, P., Yao, A., Li, Y., Hong, R., Yan, S., and Chua, T.-S.
\newblock Contrastive video question answering via video graph transformer.
\newblock \emph{arXiv preprint arXiv:2302.13668}, 2023.

\bibitem[Xu et~al.(2021{\natexlab{a}})Xu, Ghosh, Huang, Okhonko, Aghajanyan,
  Metze, Zettlemoyer, and Feichtenhofer]{xu2021videoclip}
Xu, H., Ghosh, G., Huang, P.-Y., Okhonko, D., Aghajanyan, A., Metze, F.,
  Zettlemoyer, L., and Feichtenhofer, C.
\newblock {VideoCLIP}: Contrastive pre-training for zero-shot video-text
  understanding.
\newblock In \emph{Proceedings of the Conference on Empirical Methods in
  Natural Language Processing}, 2021{\natexlab{a}}.

\bibitem[Xu et~al.(2021{\natexlab{b}})Xu, Xiong, Chen, Li, Xia, Tu, and
  Soatto]{xu2021long}
Xu, M., Xiong, Y., Chen, H., Li, X., Xia, W., Tu, Z., and Soatto, S.
\newblock Long short-term transformer for online action detection.
\newblock \emph{Advances in Neural Information Processing Systems},
  2021{\natexlab{b}}.

\bibitem[Yan et~al.(2022{\natexlab{a}})Yan, Xiong, Arnab, Lu, Zhang, Sun, and
  Schmid]{yan2022multiview}
Yan, S., Xiong, X., Arnab, A., Lu, Z., Zhang, M., Sun, C., and Schmid, C.
\newblock Multiview transformers for video recognition.
\newblock In \emph{Proceedings of the IEEE/CVF Conference on Computer Vision
  and Pattern Recognition}, 2022{\natexlab{a}}.

\bibitem[Yan et~al.(2022{\natexlab{b}})Yan, Zhu, Wang, Cao, Zhang, Ghosh, Wu,
  and Yu]{yan2022video}
Yan, S., Zhu, T., Wang, Z., Cao, Y., Zhang, M., Ghosh, S., Wu, Y., and Yu, J.
\newblock Video-text modeling with zero-shot transfer from contrastive
  captioners.
\newblock \emph{arXiv preprint arXiv:2212.04979}, 2022{\natexlab{b}}.

\bibitem[Yang et~al.(2023)Yang, Zhu, Xie, Zhang, Chen, and Li]{yang2023aim}
Yang, T., Zhu, Y., Xie, Y., Zhang, A., Chen, C., and Li, M.
\newblock {AIM: Adapting image models for efficient video action recognition}.
\newblock In \emph{International Conference on Learning Representations}, 2023.

\bibitem[Ye et~al.(2023)Ye, Xu, Yan, Xu, Qian, Zhang, and Huang]{ye2023hitea}
Ye, Q., Xu, G., Yan, M., Xu, H., Qian, Q., Zhang, J., and Huang, F.
\newblock {HiTeA}: Hierarchical temporal-aware video-language pre-training.
\newblock In \emph{Proceedings of the IEEE/CVF International Conference on
  Computer Vision}, 2023.

\bibitem[Yu et~al.(2023)Yu, Cho, Yadav, and Bansal]{yu2023self}
Yu, S., Cho, J., Yadav, P., and Bansal, M.
\newblock Self-chained image-language model for video localization and question
  answering.
\newblock In \emph{Advances in Neural Information Processing Systems}, 2023.

\bibitem[Zaheer et~al.(2020)Zaheer, Guruganesh, Dubey, Ainslie, Alberti,
  Ontanon, Pham, Ravula, Wang, Yang, et~al.]{zaheer2020big}
Zaheer, M., Guruganesh, G., Dubey, K.~A., Ainslie, J., Alberti, C., Ontanon,
  S., Pham, P., Ravula, A., Wang, Q., Yang, L., et~al.
\newblock {Big Bird}: Transformers for longer sequences.
\newblock \emph{Advances in Neural Information Processing Systems}, 2020.

\bibitem[Zeng et~al.(2022)Zeng, Attarian, Choromanski, Wong, Welker, Tombari,
  Purohit, Ryoo, Sindhwani, Lee, et~al.]{zeng2022socratic}
Zeng, A., Attarian, M., Choromanski, K.~M., Wong, A., Welker, S., Tombari, F.,
  Purohit, A., Ryoo, M.~S., Sindhwani, V., Lee, J., et~al.
\newblock Socratic models: Composing zero-shot multimodal reasoning with
  language.
\newblock In \emph{International Conference on Learning Representations}, 2022.

\bibitem[Zhang et~al.(2023)Zhang, Lu, Islam, Wang, Yu, Bansal, and
  Bertasius]{zhang2023simple}
Zhang, C., Lu, T., Islam, M.~M., Wang, Z., Yu, S., Bansal, M., and Bertasius,
  G.
\newblock {A simple LLM framework for long-range video question-answering}.
\newblock \emph{arXiv preprint arXiv:2312.17235}, 2023.

\end{thebibliography}
\bibliographystyle{icml2024}


\clearpage

\appendix
\counterwithin{figure}{section}

\section{ST-ViT and MA-ViT Pseudocode}\label{sec:app-pseudocode}

\begin{figure*}[t]
\begin{center}
\includegraphics[width=\textwidth]{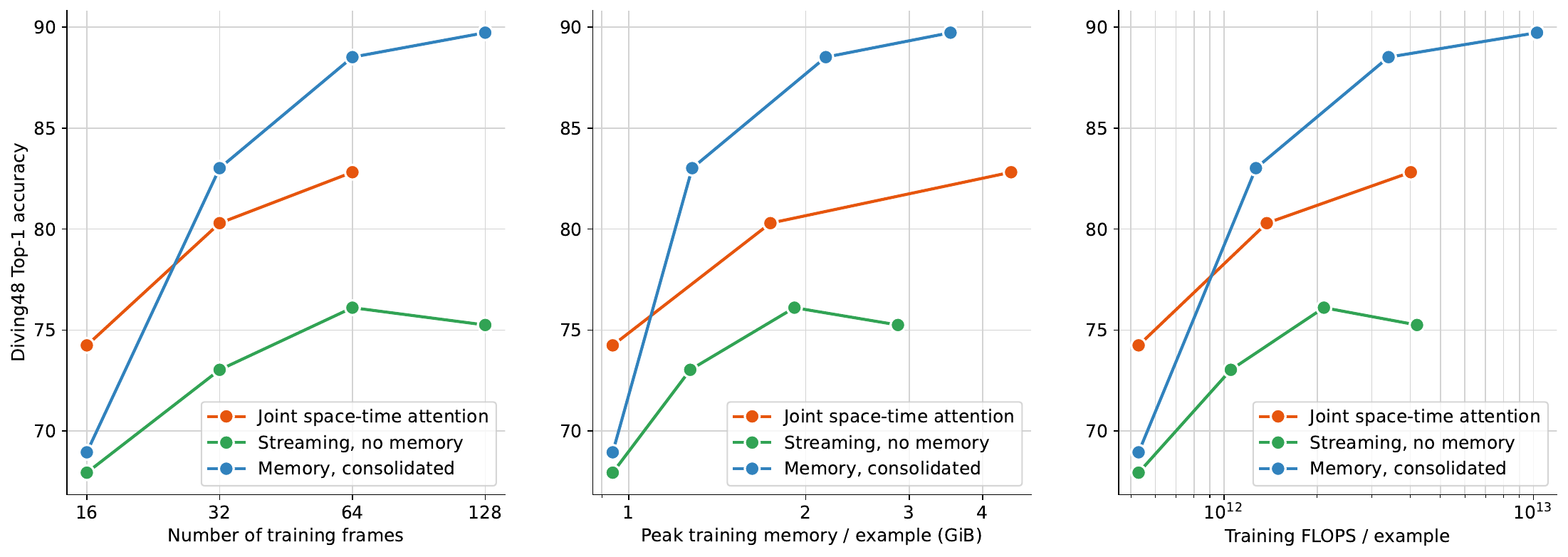}
\vspace{-2em}
\caption{\textbf{\ourmethod \ efficiently trains on long videos.} Fine-grained video understanding on Diving48 as a function of number of training frames (\textbf{left}), train-time memory consumption (\textbf{middle}), and training complexity (FLOPS, \textbf{right}), for joint space-time attention (red), memory-less streaming setting (green), and our proposed method \ourmethod \ (blue). \ourmethod \ reaches the highest accuracy with 3.5$\x$ less training memory and 3.2$\x$ fewer training FLOPS than joint space-time attention.}
\label{fig:training_efficiency}
\end{center}
\end{figure*}

We include pseudocode for streaming ViT and memory-augmented ViT in \Cref{alg:streaming_vit_algorithm} and \Cref{alg:mem_aug_vit_algorithm}.

\section{Training Details}

\subsection{Pretraining Short-Video ViViT Encoders}
\label{app:vivit_pretraining}

In this work we re-purpose a standard ViViT \cite{arnab2021vivit} encoder into \ourmethod \ and fine-tune it for long-context understanding as described in \Cref{sec:method-training}. We use a pretrained ViViT from \citet{papalampidi2023simple} (referred to as ShortViViT therein), whose training details follow those of \citet{xu2021videoclip}, and which we reproduce here for completeness. Note however that similar pretrained ViViT's are available from public repositories\footnote{\url{https://huggingface.co/docs/transformers/model_doc/vivit}}.

\noindent \textbf{Phase 1: Image-text pretraining.} A standard ViT image encoder and an associated BERT text encoder are pretrained with multimodal contrastive learning \cite{radford2021learning} on a mixture of paired image-text datasets (ALIGN,  \citet{jia2022visual}; LTIP,  \citet{alayrac2022flamingo}; and JFT, \citet{sun2017revisiting}). We utilize two variants of the underlying vision and text encoder: ViT-B with BERT-M for MC-ViT-B, as well as ViT-L and BERT-B for MC-ViT-L.

\noindent \textbf{Phase 2: Short-video pretraining.} Initializing from the phase 1 image/text encoders, \citet{papalampidi2023simple} train a ViViT video encoder and associated text encoder on short (16-frame) videos sampled from the HowTo100M \cite{miech19howto100m} and VTP \cite{alayrac2022flamingo} datasets, together with the image datasets (treated as single-frame videos) from phase 1, again using multimodal contrastive learning (Equation \ref{eq:con}). The parameters of the phase 1 ViT and phase 2 ViViT are identical except for the patch-embedding projection and positional embeddings, which they extend temporally at initialization time by simply replicating them as in \citet{arnab2021vivit}.

\subsection{Implementation Details} \label{sec:app_training_details}
We provide all experimental settings for training our \ourmethod \ model variants in Table~\ref{tab:app_training_details}, categorized by dataset. We report our experimental setup for training the large model variants, but we follow the same settings when training the base versions, except for Diving48, where we use 128 instead of 64 frames.

For training on Next-QA, we additionally use low-rank adaptation (LoRA) to avoid overfitting. Given each ViViT layer, we decompose the linear QKV input projection, the output self-attention projection, and the dense feed-forward block: 
\begin{gather}
    h = W_o x + \frac{1}{\alpha}BAx,
\end{gather}
where $W_o$ is the frozen pretrained weight matrix, $B$ and $A$ are zero-initialized low-rank matrices, $B \in \mathbb{R}^{d_m \times r}, A \in \mathbb{R}^{r \times d_m}$, $r \ll d_m$ is the rank of the decomposition matrices, and $\alpha$ is a hyperparameter for easier tuning of the model, as recommended by \cite{hulora}. We use a $r$=128, resulting in 12\% of model parameters for the large model version.

\begin{algorithm}[t!]
\caption{Streaming ViT.}
\label{alg:streaming_vit_algorithm}
\definecolor{codeblue}{rgb}{0.25,0.5,0.5}
\lstset{
  backgroundcolor=\color{ouralg!20},
  basicstyle=\fontsize{7.2pt}{7.2pt}\ttfamily\selectfont,
  columns=fullflexible,
  breaklines=true,
  captionpos=b,
  commentstyle=\fontsize{7.2pt}{7.2pt}\color{codeblue},
  keywordstyle=\fontsize{7.2pt}{7.2pt},
}
\begin{lstlisting}[language=python]
def streaming_vit(video, n_chunks, n_layers, pos_emb):
  emb = linear_proj(video) + pos_emb
  chunked_video = np.split(emb, n_chunks, axis=1)
  zs = []
  memory = None
  for z in chunked_video:
      z_norm = layer_norm(z)
      for _ in range(n_layers):
          y = self_attention(z_norm) + z
          y_norm = layer_norm(y)
          z = mlp(y_norm) + y
          zs.append(z)
  return np.concatenate(zs, axis=1)
\end{lstlisting}
\end{algorithm}

\begin{algorithm}[!t]
\caption{Memory-augmented ViT.}
\label{alg:mem_aug_vit_algorithm}
\definecolor{codeblue}{rgb}{0.25,0.5,0.5}
\lstset{
  backgroundcolor=\color{ouralg!20},
  basicstyle=\fontsize{7.2pt}{7.2pt}\ttfamily\selectfont,
  columns=fullflexible,
  breaklines=true,
  captionpos=b,
  commentstyle=\fontsize{7.2pt}{7.2pt}\color{codeblue},
  keywordstyle=\fontsize{7.2pt}{7.2pt},
}
\begin{lstlisting}[language=python]
def ma_vit(video, n_chunks, n_layers, pos_emb):
  emb = linear_proj(video) + pos_emb
  chunked_video = np.split(emb, n_chunks, axis=1)
  memory = {layer: None for layer in range(n_layers)}
  zs = []
  for z in chunked_video:
      z_norm = layer_norm(z)
      for _ in range(n_layers):
          if memory[layer] is None:
              y = self_attention(z_norm) + z
          else:
              kv = np.concatenate(z_norm, memory[layer]))
              y = cross_attention(q=z_norm, kv=kv) + z
          y_norm = layer_norm(y)
          z = mlp(y_norm) + y
          memory[layer] = memory_concatenation(memory[layer], z)
          memory[layer] = layer_norm(memory[layer])
          zs.append(z)
  return np.concatenate(zs, axis=1)
\end{lstlisting}
\end{algorithm}

\begin{algorithm}[t]
\caption{GPT-4V Zero-Shot Prompt.}
\label{alg:gpt4v-prompt}
\definecolor{codeblue}{rgb}{0.25,0.5,0.5}
\lstset{
  backgroundcolor=\color{ouralg!20},
  basicstyle=\fontsize{7.2pt}{7.2pt}\ttfamily\selectfont,
  columns=fullflexible,
  breaklines=true,
  captionpos=b,
  commentstyle=\fontsize{7.2pt}{7.2pt}\color{codeblue},
  keywordstyle=\fontsize{7.2pt}{7.2pt},
}
\begin{lstlisting}[language=python]
You are a helpful assistant, an expert in answering questions about videos. You will be given a question about a video and five possible answer options. You will be provided frames from the video, sampled evenly across the video. You are very capable, think step-by-step when answering the question.

Question: <question>

Possible answer choices:
A. <Answer Choice 1>
B. <Answer Choice 2>
C. <Answer Choice 3>
D. <Answer Choice 4>
E. <Answer Choice 5>

ONLY output A, B, C, D, or E to answer. DO NOT OUTPUT with the full answer text or any other words, output only the single letter indicating the correct choice: one of A, B, C, D, or E.

<video frames attached to the HTTP request>
\end{lstlisting}
\end{algorithm}

\begin{algorithm}[t]
\caption{GPT-4 Turbo Zero-Shot Prompt.}
\label{alg:gpt4-blind-prompt}
\definecolor{codeblue}{rgb}{0.25,0.5,0.5}
\lstset{
  backgroundcolor=\color{ouralg!20},
  basicstyle=\fontsize{7.2pt}{7.2pt}\ttfamily\selectfont,
  columns=fullflexible,
  breaklines=true,
  captionpos=b,
  commentstyle=\fontsize{7.2pt}{7.2pt}\color{codeblue},
  keywordstyle=\fontsize{7.2pt}{7.2pt},
}
\begin{lstlisting}[language=python]
You are a helpful assistant, an expert in answering questions about videos. You will be given a question about a video and five possible answer options. You will not be provided frames from the video, but still do your best to guess an answer. You are very capable, think step-by-step when answering the question.

Question: <question>

Possible answer choices:
A. <Answer Choice 1>
B. <Answer Choice 2>
C. <Answer Choice 3>
D. <Answer Choice 4>
E. <Answer Choice 5>

ONLY output A, B, C, D, or E to answer. DO NOT OUTPUT with the full answer text or any other words, output only the single letter indicating the correct choice: one of A, B, C, D, or E.
\end{lstlisting}
\end{algorithm}

\section{Additional Baselines} \label{sec:app_comparison_methods}
\vspace{-0.5em}
\textbf{Joint space-time attention with masking.} To implement joint space-time attention with masking, we follow~\citet{papalampidi2023simple} and randomly mask 25\% of the input video tokens before feeding them into the transformer. To better align training with evaluation, we perform masking both at fine-tuning and inference time. In particular, this allows the masked model to be more memory- and compute-efficient than full space-time attention. 

\textbf{Late temporal fusion.} A common approach for efficiently processing videos is applying a learnable late temporal fusion module for aggregating information over frame-level~\cite{alayrac2022flamingo,yan2022video} or short-video representations~\cite{piergiovanni2023mirasol3b}. We follow~\citet{alayrac2022flamingo} and~\citet{yan2022video} and use a light-weight Perceiver-resampler~\cite{jaegle2021perceiverio} to contextualize and compress information across segment-level information. We process the long video in 16-frame segments similarly to ST-ViT and feed all output tokens across segments into a single layer of a randomly initialized Perceiver-resampler. We use 256 latent learnable queries for learning a global compressed representation of the video.

\textbf{MeMViT.} We implement MeMViT's convolutional compression module \citep{wu2022memvit} and apply it to every video segment as a parametric alternative to  MC-ViT's consolidation mechanism. We experiment with 4$\times$ compression (i.e. kernel sizes of 1$\times$2$\times$2 and 4$\times$1$\times$1), 16$\times$ compression (kernel size 4$\times$2$\times$2) and 64$\times$ compression (kernel size 4$\times$4$\times$4). Similarly to the original paper, we find the 4$\times$2$\times$2 kernel to yield best results on the Diving48 dataset, corresponding to MC-ViT's memory segment length of $K\!=\!128$.  We also analyze the memory footprint and FLOPs of MeMViT during training and inference and we find the differences between MC-ViT and MeMViT to be almost indistinguishable, with MeMViT’s memory footprint slightly higher ($+4$\%) due to additional convolutional kernel parameters. 

\begin{table*}[t!]
\footnotesize
\centering
\caption{Training specifications for fine-tuning \ourmethod \ per dataset.}
\begin{tabular}{@{}lcccc@{}}
\toprule
& EgoSchema & Diving48 & Next-QA & Perception Test \\ \midrule
Optimizer & \multicolumn{4}{c}{AdamW} \\
Learning rate schedule & \multicolumn{4}{c}{Cosine with linear warmup} \\
Gradient clip & \multicolumn{4}{c}{2.0} \\
Linear warmup steps & \multicolumn{4}{c}{1k} \\
Frame-level resolution & \multicolumn{4}{c}{256$\times$256} \\
Frame-level cropping & \multicolumn{4}{c}{center crop} \\
Convolution kernel & \multicolumn{4}{c}{2$\times$16$\times$16} \\
Batch size & 128 & \multicolumn{3}{c}{\cellcolor{ouralg!20}256} \\
Label smoothing & 0 & \multicolumn{3}{c}{\cellcolor{ouralg!20}0.1} \\
\# memories/segment ($K$) & 128 & \multicolumn{3}{c}{\cellcolor{ouralg!20}512} \\
Frame sampling & \multicolumn{3}{c}{\cellcolor{ouralg!20}Uniform} & 4 FPS \\
Weight decay rate & 0 & 0 & \multicolumn{2}{c}{\cellcolor{ouralg!20}1e-2} \\
Base learning rate & 2e-5 & 5e-5 & \multicolumn{2}{c}{\cellcolor{ouralg!20}1e-6} \\
Training steps & 5k & 30k & \multicolumn{2}{c}{\cellcolor{ouralg!20}20k} \\

\bottomrule
\end{tabular}
\label{tab:app_training_details}
\end{table*}

\section{Evaluating Proprietary Models}
\label{app:eval_prop}

We evaluated multiple choice video question answering using the proprietary GPT-4V (``gpt-4-vision-preview") model (version as deployed on January 6 through 9th, 2023). This model has a context window of 128K tokens, with training data up to April 2023. The model was queried in a zero-shot manner, using the prompt described in Algorithm~\ref{alg:gpt4v-prompt}. Each request to the model was attached with a fixed number of frames, varying from 16 frames to 512 frames. 
We use 256$\times$256 resolution frames for all queries. Note that the token count (and corresponding cost) also increases dramatically as the number of frames accompanying the question increases: 16 frames require roughly 4.3K tokens per question/answer pair, 64 frames require 16.5K tokens, and 256 frames require 65.5K tokens. On the EgoSchema subset we found similar performance from using 16, 64, and 256 frames (63.5\%, 63.9\%, and 63.6\% accuracy respectively), hence we used 16 frames when evaluating on the full set. Frames were sampled linearly from the video, and 
we use the `auto'(-select) resolution quality setting.

For obtaining blind results, we used the ``gpt-4-1106-preview" (GPT-4 Turbo) model. The model was queried in a zero-shot manner, using the prompt described in Algorithm~\ref{alg:gpt4-blind-prompt}. Models were queried on January 17, 2023. Each request was approximately 250 tokens.

For both the blind and vision-enabled model, a small number of requests come back with a format that doesn't adhere to the A/B/C/D/E characters representing the five possible answer choices (uppercase or lowercase). For example, the model sometimes responds with ``The answer is C", or the actual text of the answer choice, instead of the corresponding answer choice letter. In those cases, we apply answer post-processing to extract out the corresponding answer choice by lowercasing the response and checking for the common regex, \texttt{the answer is (a|b|c|d|e)}. If the match fails, we compute the Levenshtein edit distance with each answer choice, choosing the lowercased answer response that most closely matches the lowercased proprietary model output.

\end{document}